\documentclass[review,12pt]{elsarticle}
\usepackage{graphicx}
\usepackage{amsmath, amssymb}
\usepackage{booktabs}
\usepackage{hyperref}
\usepackage{subcaption}
\usepackage{float}
\usepackage{tikz}
\usetikzlibrary{positioning}
\usetikzlibrary{arrows.meta}
\usepackage{pgfplots}
\pgfplotsset{compat=1.17}
\usepackage{caption}
\usepackage{tabularx} 
\usepackage{array} 
\usepackage{multirow}
\usepackage[most]{tcolorbox}
\usepackage{ltablex}
\usepackage{longtable}
\usepackage{adjustbox}
\usepackage{tcolorbox}
\usepackage{amsmath}
\usepackage{enumitem}

\journal{Neurocomputing}

\begin{document}

\begin{frontmatter}

\title{Peering Inside the Black Box: Uncovering LLM Errors in Optimization Modelling through Component-Level Evaluation}

\author[1]{Dania Refai\corref{cor1}}
\author[1,2]{Moataz Ahmed}

\address[1]{Information and Computer Science Department, KFUPM, Dhahran 31261, Saudi Arabia}

\address[2]{SDAIA-KFUPM Joint Research Center for Artificial Intelligence, King Fahd University of Petroleum \& Minerals, Dhahran, 31261, Saudi Arabia}

\cortext[cor1]{Corresponding author: Dania Refai (E-mail: Dania.Refai@hotmail.com)}

\begin{abstract}
Large language models (LLMs) are increasingly used to convert natural language descriptions into mathematical optimization formulations. Current evaluations often treat formulations as a whole, relying on coarse metrics like solution accuracy or runtime, which obscure structural or numerical errors. In this study, we present a comprehensive, component-level evaluation framework for LLM-generated formulations. Beyond the conventional optimality gap, our framework introduces metrics such as precision and recall for decision variables and constraints, constraint and objective root mean squared error (RMSE), and efficiency indicators based on token usage and latency. We evaluate GPT-5, LLaMA 3.1 Instruct, and DeepSeek Math across optimization problems of varying complexity under six prompting strategies. Results show that GPT-5 consistently outperforms other models, with chain-of-thought, self-consistency, and modular prompting proving most effective. Analysis indicates that solver performance depends primarily on high constraint recall and low constraint RMSE, which together ensure structural correctness and solution reliability. Constraint precision and decision variable metrics play secondary roles, while concise outputs enhance computational efficiency. These findings highlight three principles for NLP-to-optimization modeling: (i) Complete constraint coverage prevents violations, (ii) minimizing constraint RMSE ensures solver-level accuracy, and (iii) concise outputs improve computational efficiency. The proposed framework establishes a foundation for fine-grained, diagnostic evaluation of LLMs in optimization modeling.
\end{abstract}


\end{frontmatter}


\section{Introduction}
Large language models (LLMs) have demonstrated remarkable capabilities across diverse domains in recent years, including natural language processing (NLP), computer vision, software engineering, biomedical research, and scientific reasoning. Their ability to understand, generate, and reason over human language has enabled state-of-the-art performance in tasks such as machine translation, summarization, question answering, and code generation. Beyond these traditional applications, LLMs have emerged as promising tools for bridging the gap between natural language instructions and complex computational tasks~\cite{LLMSSurvey}, with several recent studies highlighting their use in sentiment analysis~\cite{Neuro1,Neuro3}, retrieval-augmented reasoning~\cite{Neuro2}, scientific data modeling~\cite{Neuro4}, and autonomous task benchmarking~\cite{Neuro5}.

One domain where this potential is particularly valuable is mathematical optimization. Formulating optimization problems from real-world descriptions requires substantial expertise, careful reasoning, and precise encoding of decision variables, constraints, and objective functions. The process is time-consuming and error-prone, often requiring multiple iterations between domain experts and optimization engineers. Automating this translation, from natural language descriptions to solver-ready mathematical formulations, has the potential to democratize access to optimization tools and significantly reduce modeling overhead~\cite{li2023synthesizing,ahmed2024lm4opt,huang2024optibench,abdelrahman2025teaching}.

Several recent studies have begun to explore this direction. Early approaches such as OPTIMUS~\cite{ahmaditeshnizi2024optimus}, LM4OPT~\cite{4}, and AutoFormulation~\cite{astorga2024autoformulation} demonstrated that LLMs can generate mathematical formulations directly from textual descriptions. Other works, including OptiBench~\cite{2}, ORMind~\cite{wang2025ormind}, and Chain-of-Experts~\cite{3}, have investigated advanced prompting strategies, fine-tuning, retrieval-augmented generation (RAG), and multi-agent pipelines to improve accuracy and solver readiness. Benchmarks such as NL4Opt, ComplexOR, and StarJob have also been introduced to evaluate model performance across a range of problem domains~\cite{1,2,9}. Collectively, this growing literature underscores the promise of LLMs for optimization modeling, but also reveals important gaps.

Despite these advances, several challenges remain. First, most evaluations continue to rely on proprietary models such as GPT-3.5 and GPT-4, with limited exploration of the latest generation of LLMs (e.g., GPT-5). Second, prompting strategies are often investigated in narrow contexts, with little systematic analysis of handcrafted prompts and their effectiveness across problems of varying complexity. Third, existing evaluation practices remain largely coarse-grained, focusing on global indicators such as solution accuracy, feasibility, or runtime, while overlooking structural errors in the formulation and interactions between different metrics. These limitations point to the need for a diagnostic evaluation framework that can reveal not only whether a formulation works, but also where and why errors occur.  

To address these gaps, we propose a new evaluation framework for LLM-generated optimization formulations. Unlike prior work that treats formulations as a whole, our approach evaluates them component by component, introducing metrics for decision variables (Precision and recall), constraints (Precision, recall, and constraint root mean squared error (Cons-RMSE)), and objective function root mean squared error (obj-RMSE). We complement these with solution-level quality (Optimality gap) and efficiency measures (latency and token usage). Together, these metrics provide a fine-grained view of model behavior, enabling the identification of structural weaknesses, numerical inaccuracies, and efficiency bottlenecks. We validate this framework through extensive experiments on four benchmark optimization problems of increasing complexity, applying six prompting strategies to three state-of-the-art LLMs: GPT-5, LLaMA 3.1, and DeepSeek. 

The main contributions of this paper can be summarized as follows:  
\vspace{-2mm}
\begin{itemize}
\vspace{-1mm}
    \item We design a comprehensive evaluation framework for NLP to Optimization tasks, introducing a new set of metrics that assess formulations component by component. The framework evaluates (i) Solution-level quality (Optimality gap), (ii) component-level accuracy (Precision and recall for decision variables and constraints, obj-RMSE and const-RMSE), and (iii) efficiency (Latency and token usage). By isolating individual components, our metrics precisely identify the source of errors within generated formulations, enabling a systematic analysis of the relationships between metrics and formulation correctness.  
\vspace{-3mm}
    \item We design six prompting strategies: Basic, Act-As-Expert, Chain-of-Thought, Program-of-Thought, Self-Consistency, and Modular Prompting, to investigate their effect on the quality and reliability of optimization formulations.  
\vspace{-3mm}
    \item We perform an extensive comparative evaluation of state-of-the-art LLMs, including GPT-5, DeepSeek, and LLaMA, across optimization problems of varying complexity (Easy, medium, and hard).  
\vspace{-3mm}
    \item We reveal how prompts, model reasoning, and metric interactions affect performance, identifying error patterns and trade-offs between structural accuracy, solver quality, and efficiency, providing guidance for automated optimization modeling and LLM applications in decision-making.
\end{itemize}
\vspace{-2mm}

The remainder of this paper is organized as follows. Section~\ref{sec:litreturereview} reviews related work on the use of LLMs for optimization modeling. Section~\ref{sec:evaluationMetrics} presents the evaluation framework, covering both existing metrics from the literature and our proposed extensions. Section~\ref{sec:Experimental_Setup} describes the experimental setup, including the selected optimization problems, the prompting strategies employed, and the LLMs evaluated. Section~\ref{sec:discution} reports the experimental results, beginning with problem-specific performance analyses and followed by cross-problem insights and a broader discussion of findings. Finally, Section~\ref{sec:Conclusion} concludes the paper and outlines directions for future research.

\vspace{-3mm}
\section{Literature Review}\label{sec:litreturereview}
\vspace{-1mm}
LLMs have recently shown strong potential in automating optimization problem formulation, translating natural language descriptions into mathematical models with explicit decision variables, constraints, and objective functions. Prior research has explored their reliability in generating solver-ready formulations across diverse optimization domains~\cite{li2023synthesizing,ahmed2024lm4opt,huang2024optibench,xiao2023chain}.
Existing approaches fall into two main categories. (i) In-context learning, which guides models through carefully designed prompts, zero- or few-shot examples, and advanced prompting strategies such as chain-of-thought (CoT) and tree-of-thought (ToT) reasoning~\cite{wang2024bpp,li2024nl2or,wang2024large}. This paradigm emphasizes adaptability without parameter updates and is widely used for both linear and combinatorial problems. (ii) Fine-tuning, often implemented with parameter-efficient techniques such as low-rank adaptation (LoRA), specializes models on domain-specific corpora~\cite{abgaryan2024llms,anonymous2024starjob,tang2024orlm}. Some studies further integrate fine-tuning with RAG to inject domain knowledge during inference~\cite{anonymous2024droc,zhang2024generative}.
Together, these approaches form the methodological backbone of current research on LLM-based optimization modeling.

Applications span multiple problem domains, covering both classical operations research tasks and emerging real-world challenges. Linear programming (LP) is frequently used as a benchmark because of its simplicity and established solver ecosystem~\cite{li2025abstract,deng24cafa}. More commonly, research targets combinatorial optimization, including integer and mixed-integer linear programming (ILP, MILP), which underpin applications in scheduling, routing, and resource allocation~\cite{xiao2023chain,abgaryan2024llms,anonymous2024starjob,wang2024leveraging}. A smaller but growing body of work explores hybrid formulations that combine linear and combinatorial elements~\cite{tang2024orlm,jiang2024llmopt}. Overall, combinatorial optimization dominates the literature, providing a rich testbed for evaluating LLM reasoning and formulation capabilities~\cite{huang2024optibench,zhou2024llmsolver}.

To benchmark LLM performance, researchers have introduced diverse datasets spanning synthetic and real-world optimization problems. Commonly used benchmarks include NL4Opt, LP word problems (LPWP), Mamo, IndustryOR, NLP4LP, ComplexOR, and OptiBench, as well as specialized corpora such as StarJob, ReSocratic, and Cycle Share~\cite{li2023synthesizing,ahmed2024lm4opt,huang2024optibench,abgaryan2024llms,jiang2024llmopt,jiao2024city}. These datasets vary in scale and complexity, from simple word problems to industrial-grade tasks, supporting systematic evaluation of LLM formulation capability. Among them, NL4Opt remains the most widely adopted due to its accessibility and diversity~\cite{li2023synthesizing,ahmed2024lm4opt,jiang2024llmopt}.
For solver evaluation, generated formulations are typically tested using CPLEX, SCIP, COPT, PuLP, and OR-Tools, with Gurobi emerging as the dominant choice owing to its robustness and broad support for linear and combinatorial problems~\cite{huang2024optibench,xiao2023chain,wang2024leveraging,zhang2025or}.

A broad range of LLMs has been applied to optimization tasks, spanning proprietary and open-source models. The most frequently studied include GPT-4, GPT-3.5, and LLaMA, which dominate the literature due to their strong reasoning ability and accessibility~\cite{li2023synthesizing,huang2024optibench,abgaryan2024llms,anonymous2024starjob}. Other notable models, Mistral, DeepSeek, Qwen, PaLM, Zephyr, Phi, Mixtral, Falcon, and Claude, have been explored in specialized contexts or paired with advanced prompting strategies~\cite{tang2024orlm,jiang2024llmopt,chen2025solver,huang2025llms}.
Despite increasing interest in open-source alternatives, proprietary GPT models continue to dominate comparative studies~\cite{wang2025ormind,zhang2025or}. This trend highlights their superior performance yet raises reproducibility and transparency concerns, prompting a growing shift toward open-source and fine-tuned variants~\cite{liu2024variable,mostajabdaveh2024optimization}.

The effectiveness of LLMs in optimization formulation largely depends on prompting and adaptation strategies. Most studies employ in-context learning, using carefully designed prompts, zero- or few-shot examples, and reasoning strategies such as CoT and ToT to guide model outputs~\cite{wang2024bpp,li2024nl2or,wang2024large}. Others leverage self-consistency and multi-agent prompting, where multiple LLMs collaborate to refine or validate formulations~\cite{xiao2023chain,mostajabdaveh2024optimization,nammouchi2024towards}.
Beyond prompting, RAG has been explored to inject domain-specific knowledge~\cite{anonymous2024droc,zhang2024generative}, while fine-tuning with parameter-efficient methods such as LoRA adapts models for optimization modeling tasks~\cite{abgaryan2024llms,anonymous2024starjob,tang2024orlm}.

Evaluation practices in recent literature employ diverse metrics to assess LLM-generated optimization formulations. Common measures include solution quality (e.g., optimality gap), surface-form accuracy (e.g., F1), buildability and robustness (e.g., compilation accuracy), efficiency (e.g., solving time), mathematical fidelity (e.g., integrity gap), domain utility (e.g., utility improvement), and human-centered evaluation. Although widely adopted, these metrics have notable limitations when used in isolation, as they often fail to capture structural correctness or reveal the sources of formulation errors. Our study directly addresses these gaps; hence, a detailed discussion of existing metrics, their shortcomings, and our proposed evaluation framework is provided in Section~\ref{sec:evaluationMetrics}. 

In summary, while prior studies have demonstrated the ability of LLMs to generate optimization formulations, several important gaps remain. First, most evaluations continue to rely on proprietary models such as GPT-3.5 and GPT-4, with no studies yet exploring the latest generation of models, such as GPT-5 or systematically comparing them with strong open-source alternatives like Llama and DeepSeek. Second, prompting strategies are often tested in narrow settings, with little systematic analysis of handcrafted prompts or their behavior across optimization problems of different complexity levels. Third, while a wide range of evaluation metrics has been introduced, these metrics remain largely coarse-grained, focusing on overall solution accuracy, feasibility, or runtime, without diagnosing where errors occur within the formulation or how different metrics interact. These limitations highlight the need for a comprehensive evaluation framework that can assess optimization formulations at a finer granularity, account for both structural and numerical correctness, and evaluate models more rigorously across prompts and problems of varying complexity. Addressing this gap is the central objective of our work, and we provide a detailed discussion of evaluation metrics, their shortcomings, and our proposed framework in Section~\ref{sec:evaluationMetrics}.

\vspace{-3mm}
\section{Evaluation Metrics}\label{sec:evaluationMetrics}

The evaluation of LLM-generated optimization formulations is a challenging task because a formulation is not a monolithic object but a structured composition of \textbf{decision variables}, \textbf{constraints}, and an \textbf{objective function}. Each of these components must be generated correctly for the overall formulation to be valid and useful. Accordingly, evaluation metrics should ideally capture correctness at both the \emph{solution level} and the \emph{component level}. In this section, we review existing metrics, highlight their limitations, and present our proposed framework that addresses these gaps.

\vspace{-3mm}
\subsection{Metrics in the Literature}
Evaluation practices in the literature employ a diverse set of metrics designed to assess both solution quality and structural correctness. Solution-based metrics quantify the degree to which LLM-generated solutions match reference results or solver outputs. Symbolic-level metrics evaluate syntactic accuracy, while robustness measures examine the model’s reliability under varying conditions. Additional metrics focus on efficiency (e.g., runtime, convergence), mathematical fidelity (e.g., integrity gap, semantic similarity), and domain-specific utility (e.g., hypervolume, IGD). Complementing these quantitative measures, human-centered evaluations provide expert qualitative assessments of solution clarity, interpretability, and practical usefulness~\cite{abdelrahman2025teaching}. A concise summary of all metrics, along with their definitions and representative studies, is presented in Table~\ref{tab:eval_metrics}.

\begin{table*}[!h]
\centering
\caption{Summary of evaluation metrics for LLM-generated optimization formulations.}
\label{tab:eval_metrics}

\renewcommand{\arraystretch}{0.9} 
\setlength{\tabcolsep}{1.5pt}    
\scriptsize                      

\begin{adjustbox}{max width=\textwidth, max totalheight=\textheight, keepaspectratio}
\begin{tabular}{|>{\centering\arraybackslash}m{3.6cm}
                ||>{\centering\arraybackslash}m{3.0cm}
                |>{\arraybackslash}m{7.5cm}
                |>{\centering\arraybackslash}m{3.0cm}|}
\hline
\textbf{Category} & \textbf{Metric} & \textbf{Definition} & \textbf{Representative Studies} \\
\hline \hline

\multirow{6}{*}{\centering Solution quality} 
 & Solution accuracy & Deviation of obtained solution from the best-known or optimal value. & \cite{huang2024optibench,wang2024bpp,li2023synthesizing,ahmed2024lm4opt,tang2024orlm,jiang2024llmopt,chen2025solver,amarasinghe2023aicopilotbusinessoptimisationframework,anonymous2024droc,ahmaditeshnizi2024optimus,xiao2023chain,chen2024diagnosing,anonymous2024optibench,yang2024large,wang2024large,astorga2024autoformulation,jiao2024city,wang2025ormind,huang2025llms,nammouchi2024towards,mostajabdaveh2024optimization,li2023large,tang2024orlm,deng24cafa,zhang2025or,li2025abstract,zhang2025decision} \\
\cline{2-4}
 & Optimality gap & Relative difference between achieved objective and known optimum. & \cite{abgaryan2024llms,anonymous2024starjob,zhou2024llmsolver,wang2024leveraging,anonymous2024droc,sun2024generative,hao2024planning} \\
\cline{2-4}
 & Relative regret & Loss in performance vs. the optimal solution. & \cite{zhang2024generative} \\
\cline{2-4}
 & Avg. improvement ratio & Comparison vs. heuristic baselines; $<1$ = better than baseline. & \cite{zhou2024llmsolver} \\
\hline

Surface-form accuracy & Precision, Recall, F1-score & Token-level comparison to reference formulations. & \cite{kadiouglu2024ner4opt} \\
\hline

\multirow{3}{*}{\centering Buildability \& Robustness}
 & Compilation accuracy & Whether code parses successfully and solver accepts it without syntax or schema errors. & \cite{xiao2023chain,kadiouglu2024ner4opt} \\
\cline{2-4}
 & Execution rate & Proportion of compilable runs that complete successfully without runtime errors. & \cite{wang2025ormind,jiang2024llmopt,chen2025solver,zhang2025or} \\
\hline

\multirow{4}{*}{\centering Efficiency \& Search}
 & Solving / runtime & Average time for solver to produce results. & \cite{luzzi2025chatgpt,jiang2024llmopt,xiao2023chain,zhang2024generative,nammouchi2024towards} \\
\cline{2-4}
 & Convergence rate & Time to reach within 1\% of best-known solution. & \cite{zhang2024generative} \\
\cline{2-4}
 & Valid@k (time $t$) & Fraction where at least one of top-$k$ solutions is valid within time limit. & \cite{li2024nl2or} \\
\hline

\multirow{3}{*}{\centering Domain utility}
 & Utility improvement & Gain in native units (e.g., Mbps, \$, min) vs. baseline. & \cite{gemp2024steering,zhang2024generative} \\
\cline{2-4}
 & Hypervolume & Extent of objective space covered by produced Pareto set. & \cite{yao2024multi} \\
\cline{2-4}
 & Inv. gen. distance & Distance of produced Pareto set to the reference front. & \cite{yao2024multi} \\
\hline

\multirow{3}{*}{\centering Mathematical fidelity}
 & Integrity gap & Structural mismatch in variables, constraints, and associations; $0$ = perfect match. & \cite{li2024towards} \\
\cline{2-4}
 & Semantic similarity & Equivalence of meaning even if expressed differently; measured via embeddings or LLM-judge. & \cite{liu2024variable} \\
\hline

Human-centered evaluation & Expert assessment & Qualitative evaluation of clarity, maintainability, and practical usefulness. & \cite{nammouchi2024towards} \\
\hline

\end{tabular}
\end{adjustbox}
\end{table*}


\subsection{Limitations of Existing Metrics}
Although the literature reports a broad spectrum of metrics, these measures remain limited when applied in isolation. In particular:
\vspace{-3mm}
\begin{itemize}
    \item \textbf{Solution-based metrics can be misleading:} Metrics such as the optimality gap may suggest high solution quality even when the generated formulation is structurally flawed. For instance, an LLM-generated model may achieve an optimality gap close to $0$ while omitting critical constraints, as conceptually illustrated in Fig.~\ref{fig:illustrative_example}. To demonstrate this issue with a real-world example, we evaluated an LLM on the aircraft assignment problem from the ComplexOR dataset~\cite{COMPLEXOR}. As shown in Fig.~\ref{fig:real_constraint_example}, the LLM achieved perfect solution quality (optimality gap = 0) despite identifying only $50$\% of the ground truth constraints correctly; it missed the binary and capacity constraints while incorrectly adding a non-negativity constraint. In such cases, the optimality gap fails to reflect the unreliability of the formulation. This highlights the need for component-level metrics that evaluate not only how well a model solves, but also what it generates in terms of decision variables, constraints, and objectives.

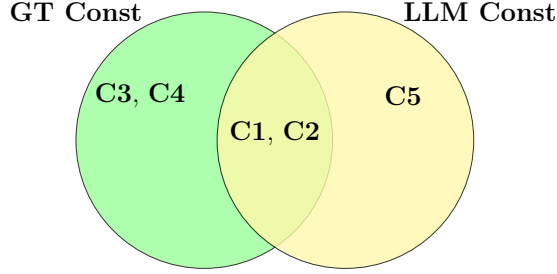
\begin{figure}[htbp]

\centering
\resizebox{0.65\linewidth}{!}{%
\begin{tikzpicture}
  \begin{scope}
    \clip (0,0) circle (2);
    \fill[yellow!60] (2.2,0) circle (2); 
  \end{scope}
  \draw[fill=green!40,opacity=0.8] (0,0) circle (2); 
  \draw[fill=yellow!40,opacity=0.8] (2.2,0) circle (2); 

  \node at (-2.0,2.0) {\textbf{GT Const}};
  \node at (4.3,2.0) {\textbf{LLM Const}}; 

  \node at (-1.0,0.7) {\textbf{C3}, \textbf{C4}}; 
  \node at (3.1,0.7) {\textbf{C5}};               
  \node at (1.1,0.1) {\textbf{C1}, \textbf{C2}};  

  \node at (1.2,-2.8) {\textbf{Precision} = 2/3, \textbf{Recall} = 2/4, \textbf{Optimality gap} $\approx 0$};
\end{tikzpicture}
}
\caption{LLM achieves a near-zero optimality gap but misses constraints, which shows why component-level metrics are necessary. GT Const = Ground Truth Constraints, LLM Const = LLM-generated Constraints.}
\label{fig:illustrative_example}
\end{figure}

\begin{figure}[htbp]
\centering
\resizebox{1\linewidth}{!}{%
\begin{tikzpicture}[
    constraint/.style={rectangle, draw, rounded corners, minimum width=6cm, minimum height=0.5cm, align=left, font=\footnotesize},
    header/.style={font=\bfseries\footnotesize},
    status/.style={font=\tiny, align=center}
]

\node[header] (gt_header) at (-3.2,1.8) {Ground Truth Constraints (4 total)};
\node[header] (llm_header) at (3.2,1.8) {LLM Constraints (3 total)};

\node[constraint, fill=green!20] (gt1) at (-3.2,1.23) {\textbf{Availability:} $\sum_j x_{ij} \leq A_i$};
\node[constraint, fill=green!20] (gt2) at (-3.2,0.4) {\textbf{Demand:} $\sum_i x_{ij} \geq D_j$};
\node[constraint, fill=red!20] (gt3) at (-3.2,-0.4) {\textbf{Binary:} $x_{ij} \in \{0,1\}$};
\node[constraint, fill=red!20] (gt4) at (-3.2,-1.2) {\textbf{Capacity:} $\sum_i x_{ij} \cdot \text{Cap}_{ij} \geq D_j$};

\node[constraint, fill=green!20] (llm1) at (3.2,1.23) {\textbf{Availability:} $\sum_j x_{ij} \leq A_i$};
\node[status] at (5.8,1.25) {\textcolor{green}{TP}};

\node[constraint, fill=green!20] (llm2) at (3.2,0.4) {\textbf{Demand:} $\sum_i x_{ij} \geq D_j$};
\node[status] at (5.8,0.4) {\textcolor{green}{TP}};

\node[constraint, fill=orange!20] (llm3) at (3.2,-0.4) {\textbf{Non-negativity:} $x_{ij} \geq 0$};
\node[status] at (5.8,-0.4) {\textcolor{orange}{FP}};

\node[constraint, draw=red, dashed, fill=red!5] (llm4) at (3.2,-1.2) {\textit{(Missing constraint)}};
\node[status] at (5.8,-1.18) {\textcolor{red}{FN}};

\node[status] at (-5.9,1.25) {\textcolor{green}{C}};
\node[status] at (-5.9,0.4) {\textcolor{green}{C}};
\node[status] at (-5.9,-0.4) {\textcolor{red}{M}};
\node[status] at (-5.9,-1.18) {\textcolor{red}{M}};

\node[draw, rounded corners, fill=white, align=center, font=\footnotesize, minimum width=8.8cm, minimum height=0.6cm] (perf_title) at (0,-2.0) {
    \textbf{Performance Summary:}
};

\node[draw, rounded corners, fill=white, align=center, font=\footnotesize, minimum width=8cm, minimum height=0.8cm] (perf_metrics) at (0,-3.2) {
    \begin{tabular}{llll}
        Precision & $\frac{2}{2+1}=0.67$ & Recall & $\frac{2}{2+2}=0.50$ \\
        Optimality Gap & 0.00 & & \\
    \end{tabular}
};

\node[draw, rounded corners, fill=gray!10, align=center, font=\tiny, minimum width=8.8cm, minimum height=0.5cm] (legend) at (0,-4.35) {
    \textcolor{green}{TP=Correct constraint} | 
    \textcolor{red}{FN=Missed constraint} | 
    \textcolor{orange}{FP=Wrong constraint}
};

\end{tikzpicture}
}
\caption{Real experiment showing constraint identification vs. solution quality. The LLM correctly detected availability and demand constraints (TP = 2), missed binary and capacity ones (FN = 2), and added a non-negativity constraint incorrectly (FP = 1). Despite these errors, the solution quality remained perfect.}
\label{fig:real_constraint_example}
\end{figure}

    \item \textbf{Symbolic accuracy is restricted to token-level comparisons:} Metrics such as precision, recall, and F1, used in studies like Ner4Opt~\cite{kadiouglu2024ner4opt}, treat the entire optimization formulation as a sequence of tokens rather than evaluating individual components like variables, constraints, or objectives. This means they measure how closely the generated tokens match the reference, without indicating whether key elements are correct or missing. As a result, a model may achieve high token-level scores while producing structurally incomplete or incorrect formulations. Moreover, token-level F1 is highly sensitive to superficial differences, such as variable naming, indexing, or formatting, potentially underestimating semantic equivalence between mathematically identical formulations.
\vspace{-3mm}
    \item \textbf{Numerical fidelity of objectives is overlooked:} Even when an LLM generates an objective function that appears syntactically correct, small coefficient errors can alter trade-offs or decision priorities. Prior studies rarely quantify this functional closeness (e.g., via \emph{obj-RMSE} over sampled inputs), making it difficult to detect numerically inaccurate but superficially plausible objectives.
\vspace{-3mm}
    \item \textbf{Constraint behavior is not assessed numerically:} Feasibility is often treated as a binary outcome (feasible/infeasible) or inferred indirectly from solution quality. However, this ignores how closely generated constraints \emph{behave} relative to their ground-truth counterparts under identical variable assignments. A function-level measure such as \emph{const-RMSE} can reveal coefficient errors, sign mistakes, or misplaced variable terms that binary feasibility checks miss.
\vspace{-3mm}
    \item \textbf{Efficiency measures are incomplete: } Existing works typically focus on solver runtime while ignoring the computational cost of using LLMs themselves. Metrics such as \textbf{latency} and \textbf{token usage} are crucial for assessing the practical deployment of LLMs in optimization pipelines.
\vspace{-3mm}
    \item \textbf{Interdependence between metrics is unexplored:} No prior study has systematically analyzed the relationships between metrics, making it difficult to understand trade-offs among structural correctness, numerical fidelity, and efficiency.
\vspace{-3mm}
    \item \textbf{Formulation components are not evaluated separately:} Critically, no existing work assesses decision variables, constraints, and objectives as distinct components of the formulation. Without such granularity, diagnostic insights into model weaknesses remain limited.
\end{itemize}
\vspace{-3mm}
These limitations motivate the introduction of component- and function-level metrics that capture both structural correctness and numerical fidelity, forming the foundation of our proposed comprehensive evaluation framework.

\vspace{-3mm}
\subsection{Proposed Evaluation Framework for LLM-Generated Formulations}\label{Evaluation}

To systematically evaluate the performance of LLMs in generating optimization formulations, we introduce a set of metrics designed to capture both symbolic correctness and functional fidelity. These metrics extend beyond traditional solution-based measures by explicitly assessing the structural, numerical, and efficiency aspects of generated formulations. 

First, \textbf{constraint precision and recall} quantify how accurately the model identifies and reconstructs the symbolic structure of constraints. 
Second, \textbf{decision variable precision and recall} evaluate the model’s ability to correctly extract and represent the relevant decision variables. 
Third, the \textbf{optimality gap} measures the difference in objective value between the solution obtained from the LLM-generated formulation and that of the ground-truth formulation, thereby reflecting overall solution quality. 
Fourth, the \textbf{objective function RMSE} assesses numerical fidelity by computing the deviation between the generated and reference objective functions across sampled inputs. 
Fifth, the \textbf{constraint RMSE} evaluates the functional accuracy of generated constraints by measuring their numerical error relative to the ground truth. 
Finally, a set of \textbf{efficiency metrics}, including response latency and token usage, captures the computational cost and practicality of deploying LLMs in optimization modeling. In the following subsections, we provide formal definitions of each metric.

\vspace{-3mm}
\subsubsection{Constraints Precision and Recall}\label{constraints}

We propose the constraints precision and recall metrics to evaluate the structural accuracy of constraint generation in LLM-produced optimization formulations. Unlike prior evaluation approaches that assess token-level similarity or overall formulation overlap, this metric specifically measures how accurately an LLM reproduces the set of constraints that define the feasible region of the optimization problem. Accurately capturing these constraints is essential, as missing or incorrect constraints can lead to infeasible or suboptimal solutions, even when the objective function is correctly specified.

Let true positives ($\text{TP}_{\text{constraints}}$) represent the constraints correctly generated by the LLM that also appear in the ground-truth formulation. False positives ($\text{FP}_{\text{constraints}}$) are constraints generated by the LLM that do not exist in the ground truth, while false negatives ($\text{FN}_{\text{constraints}}$) are constraints present in the ground-truth formulation but missing from the LLM’s output. Based on these definitions, precision and recall for constraints are computed as follows: 
\begin{equation}
\text{Precision}_{\text{constraints}} = \frac{\text{TP}_{\text{constraints}}}{\text{TP}_{\text{constraints}} + \text{FP}_{\text{constraints}}} \label{eq:precision_constraints} 
\end{equation} 
\vspace{-5mm}
\begin{equation} 
\text{Recall}_{\text{constraints}} = \frac{\text{TP}_{\text{constraints}}}{\text{TP}_{\text{constraints}} + \text{FN}_{\text{constraints}}} \label{eq:recall_constraints} 
\end{equation} 

Here, $\text{Precision}_{\text{constraints}}$ quantifies the proportion of generated constraints that are correct, while $\text{Recall}_{\text{constraints}}$ measures the proportion of ground-truth constraints that are successfully recovered by the LLM. Together, these metrics offer complementary perspectives on the structural accuracy of constraint generation.

\subsubsection{Decision Variables Precision and Recall}\label{decision_variables}

We introduce the decision variables precision and recall metrics to evaluate how accurately an LLM identifies and reproduces the decision variables that define the solution space of an optimization problem. Decision variables are fundamental to the structure of any formulation, as they represent the quantities the model must determine to achieve an optimal solution. Errors in identifying these variables, such as missing, redundant, or misclassified ones, can compromise the entire optimization process, even if constraints and objectives are correctly generated.

In this context, true positives ($\text{TP}_{\text{variables}}$) refer to decision variables correctly generated by the LLM that also appear in the ground-truth formulation. False positives ($\text{FP}_{\text{variables}}$) are variables introduced by the LLM that are not present in the ground truth, while false negatives ($\text{FN}_{\text{variables}}$) are ground-truth decision variables omitted from the LLM’s output. Precision and recall for decision variables are defined as follows:
\begin{equation}
\text{Precision}_{\text{variables}} = \frac{\text{TP}_{\text{variables}}}{\text{TP}_{\text{variables}} + \text{FP}_{\text{variables}}} 
\label{eq:precision_variables}
\end{equation}
\vspace{-5mm}
\begin{equation}
\text{Recall}_{\text{variables}} = \frac{\text{TP}_{\text{variables}}}{\text{TP}_{\text{variables}} + \text{FN}_{\text{variables}}} 
\label{eq:recall_variables}
\end{equation}

Here, $\text{Precision}_{\text{variables}}$ reflects the proportion of generated variables that are relevant, indicating how well the model avoids introducing incorrect variables. $\text{Recall}_{\text{variables}}$, on the other hand, captures the model’s ability to identify all relevant decision variables from the ground truth. Together, these metrics provide a comprehensive view of the model’s performance in recognizing the structural components of the decision space.

\vspace{-3mm}
\subsubsection{Optimality Gap (Adopted from Prior Literature)}\label{optimality_gap}

The optimality gap is a standard metric for assessing how closely a solution approximates the true optimum. Here, it evaluates the numerical quality of LLM-generated formulations by measuring the relative deviation between the LLM’s objective value and the known optimum; a smaller gap indicates higher solution quality.

To compute this metric, the LLM-generated formulation is first converted into an executable optimization model and solved using the Gurobi solver~\cite{gurobi_solver}. The resulting objective value is then compared against the ground-truth optimal value. The optimality gap is defined as:
\begin{equation}
\small
\text{Optimality gap} = \frac{|\text{Optimal value} - \text{LLM objective value}|}{|\text{Optimal value}|}
\label{eq:optimality_gap}
\end{equation}

This metric provides a direct and interpretable measure of solution quality, with values approaching zero indicating that the LLM effectively reproduces the optimal behavior of the ground-truth formulation.


\subsubsection{RMSE for Objective Function Comparison}\label{rmse}

We are proposing the objective function root mean squared error (obj-RMSE) to evaluate the functional accuracy of the objective functions generated by the LLMs. Specifically, this metric aims to capture how numerically close the LLM-generated objective values are to the ground-truth values across a range of problem instances, regardless of the underlying symbolic structure. This allows us to assess whether the generated objectives behave similarly to the true ones when evaluated over concrete inputs. A lower obj-RMSE indicates stronger agreement with the ground truth, while a higher obj-RMSE reflects greater deviation and reduced fidelity.

Let $\text{True objective}_i$ denote the optimal objective value for the $i$-th instance, $\text{LLM objective}_i$ the corresponding value produced by the LLM, and $n$ the total number of evaluated instances. The Obj-RMSE is calculated as:
\begin{equation}
\small
\text{Obj-RMSE} = \sqrt{\frac{1}{n} \sum_{i=1}^{n} \left(\text{True objective}_i - \text{LLM objective}_i\right)^2}
\label{eq:rmse}
\end{equation}

This metric provides a direct, quantitative measure of the extent to which the LLM's objective function approximates the numerical behavior of the true formulation. 

\vspace{-3mm}
\subsubsection{RMSE for Constraint Behavior Comparison}\label{cons-rmse}

The proposed constraint root mean squared error (Cons-RMSE) is used to evaluate how well an LLM’s generated constraints replicate the functional behavior of the ground-truth formulation. Unlike symbolic metrics such as constraint precision and recall, which assess structural or syntactic correctness, Cons-RMSE captures the numerical alignment of matched constraints when evaluated on identical input values.

The core aim of Cons-RMSE is to answer the following question: \textit{If we plug the same decision variable values into both the LLM-generated constraints and the ground-truth constraints, do they behave the same numerically?} A low Cons-RMSE indicates strong behavioral fidelity, suggesting that the LLM’s constraints closely mimic the ground truth in practice. Conversely, a high Cons-RMSE implies that, even if a constraint appears structurally correct, it behaves differently, for example, due to incorrect coefficients or misapplied variable terms.

To compute this metric, we first calculate the const-RMSE individually for each matched constraint by comparing its evaluations over a fixed set of decision variable samples with those of the corresponding ground-truth constraint. We then take the average const-RMSE across all matched constraints to obtain the final Cons-RMSE score for the formulation. This averaging approach allows us to assess the overall behavioral correctness of the model’s output without being skewed by any single constraint.

Importantly, Cons-RMSE is computed only over the subset of constraints that are considered matched between the LLM-generated and ground-truth formulations. Constraints that are hallucinated, omitted, or structurally incompatible are excluded from this analysis. This design ensures that Cons-RMSE focuses solely on the correctness of the constraints the LLM attempted to reproduce, complementing symbolic metrics: Precision penalizes extraneous constraints, recall penalizes missing ones, while Cons-RMSE evaluates the functional accuracy of those that are present and matched.

Let \( x^{(i)} \) denote the $i$-th decision variable sample, \( f_k^{\text{GT}}(x^{(i)}) \) the evaluation of the $k$-th ground-truth constraint on that input, and \( f_k^{\text{LLM}}(x^{(i)}) \) the corresponding evaluation of the LLM-generated constraint. Let \( \mathcal{C} \) be the set of matched constraints and \( n \) the number of input samples. The Cons-RMSE is computed as:
\begin{equation}
\small
\text{Cons-RMSE} = \sqrt{
\frac{1}{|\mathcal{C}| \cdot n}
\sum_{k \in \mathcal{C}} \sum_{i=1}^{n}
\left( f_k^{\text{GT}}(x^{(i)}) - f_k^{\text{LLM}}(x^{(i)}) \right)^2}
\label{eq:cons-rmse}
\end{equation}

In the experimentation, to ensure fairness and consistency across all evaluations, we use a fixed set of \( n = 100 \) randomly generated input samples for each problem. These samples are drawn from valid input domains tailored to each task (e.g., continuous time intervals for scheduling problems or binary vectors for combinatorial tasks like knapsack) using a fixed random seed. This controlled sampling process eliminates randomness as a confounding factor, enabling a reliable assessment of functional correctness.


\vspace{-3mm}
\subsubsection{Efficiency}\label{efficiency}

Efficiency evaluates the practical performance of LLMs in optimization tasks, focusing on computational resources and responsiveness. It is measured by input tokens (Total tokens given to the model, with fewer indicating efficient problem interpretation), output tokens (Tokens generated, with fewer reflecting concise formulations), and latency (Time to generate a response, with lower values indicating faster performance). These metrics capture the trade-off between computational cost and solution quality.

\begin{figure*}[htbp]
\centering
\resizebox{1.06\linewidth}{!}{%
\scalebox{0.5}{
\begin{tikzpicture}[node distance=0.2cm, >=stealth, thick]

\node[draw, double, double distance=3pt, fill=cyan!30, rounded corners, minimum width=6cm, minimum height=1.5cm, align=center, line width=1.5pt] (central) 
{\textbf{Evaluation Metrics for LLM-based Optimization Formulations}};

\node[draw, double, double distance=3pt, fill=yellow!50, rounded corners, below left=1cm and 0.5cm of central, minimum width=7cm, minimum height=1.5cm, align=center,line width=1.5pt] (literature) {\textbf{Literature Metrics}};

\node[draw, fill=yellow!20, rounded corners, below=of literature, minimum width=7cm, align=center, line width=1.5pt] (sol) {\textbf{Solution Quality} \\ \scriptsize (Accuracy, Optimality Gap, AIR, Regret)}; 
\node[draw, fill=yellow!20, rounded corners, below=of sol, minimum width=7cm, align=center, line width=1.5pt] (surf) {\textbf{Surface-form Accuracy} \\ \scriptsize (Precision, Recall, F1)};
\node[draw, fill=yellow!20, rounded corners, below=of surf, minimum width=7cm, align=center, line width=1.5pt] (rob) {\textbf{Buildability \& Robustness} \\ \scriptsize (Compilation, Execution)};
\node[draw, fill=yellow!20, rounded corners, below=of rob, minimum width=7cm, align=center, line width=1.5pt] (eff) {\textbf{Efficiency \& Search} \\ \scriptsize (Runtime, Convergence, Valid@k)};
\node[draw, fill=yellow!20, rounded corners, below=of eff, minimum width=7cm, align=center, line width=1.5pt] (fid) {\textbf{Mathematical Fidelity} \\ \scriptsize (Integrity Gap, Semantic Sim.)};
\node[draw, fill=yellow!20, rounded corners, below=of fid, minimum width=7cm, align=center, line width=1.5pt] (dom) {\textbf{Domain Utility} \\ \scriptsize (Utility, HV, IGD)};
\node[draw, fill=yellow!20, rounded corners, below=of dom, minimum width=7cm, align=center, line width=1.5pt] (hum) {\textbf{Human Evaluation} \\ \scriptsize (Expert Assessment)};

\node[draw, double, double distance=3pt, fill=orange!50, rounded corners, below right=1.0cm and 0.5cm of central, minimum width=7cm, minimum height=1.5cm, align=center, line width=1.5pt] (ourwork) {\textbf{Our Work}};

\node[draw, fill=orange!30, rounded corners, below=of ourwork, minimum width=7cm, align=center] (struct)
{\textbf{Structural Precision \& Recall} \\ \scriptsize (Decision Variables, Constraints)};
\node[draw, fill=orange!30, rounded corners, below=of struct, minimum width=7cm, align=center, line width=1.5pt] (num)
{\textbf{Numerical Accuracy} \\ \scriptsize (RMSE for Objectives, Constraints)};
\node[draw, fill=yellow!20, rounded corners, below=of num, minimum width=7cm, align=center, line width=1.5pt] (sqgap)
{\textbf{Solution Quality} \\ \scriptsize (Optimality Gap)};
\node[draw, fill=orange!30, rounded corners, below=of sqgap, minimum width=7cm, align=center, line width=1.5pt] (exteff)
{\textbf{Extended Efficiency} \\ \scriptsize (Latency, Token Usage)};

\draw[-{Stealth[length=10pt,width=12pt]}, line width=1.5pt] 
    (central.south) to[out=250,in=80] (literature.north);
\draw[-{Stealth[length=10pt,width=12pt]}, line width=1.5pt] 
    (central.south) to[out=290,in=100] (ourwork.north);

\end{tikzpicture}
}
}
\caption{Taxonomy of evaluation metrics.}
\label{fig:taxonomy_gap}
\end{figure*}
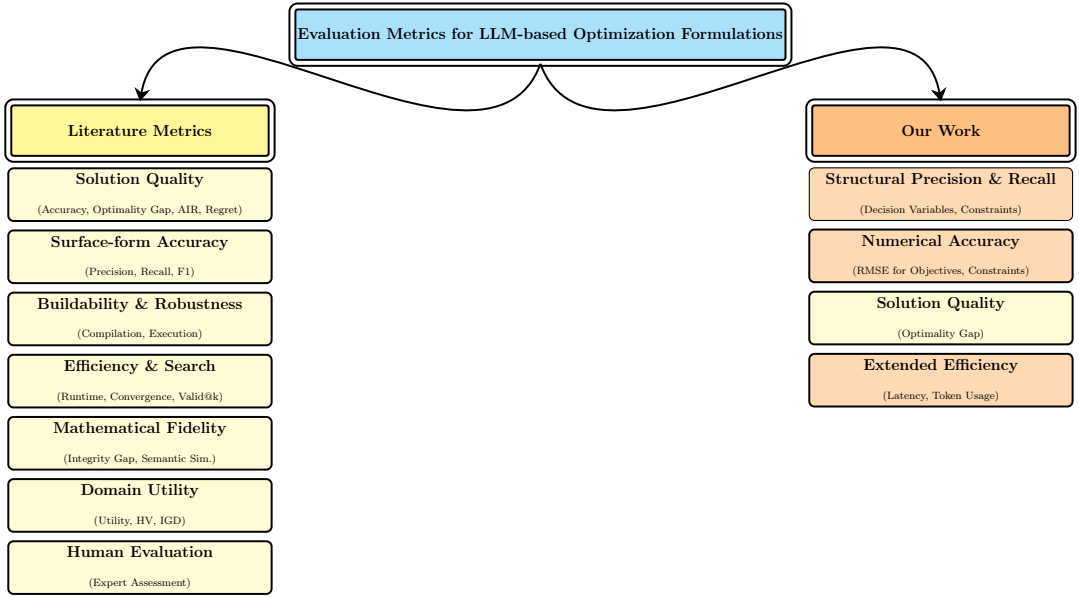

Fig.~\ref{fig:taxonomy_gap} presents a taxonomy of evaluation metrics, summarizing both those reported in the literature and the new metrics proposed in this work.

\vspace{-3mm}
\section{Experiment Setup}\label{sec:Experimental_Setup}

\begin{figure*}[htbp]
    \centering
    \includegraphics[width=\columnwidth]{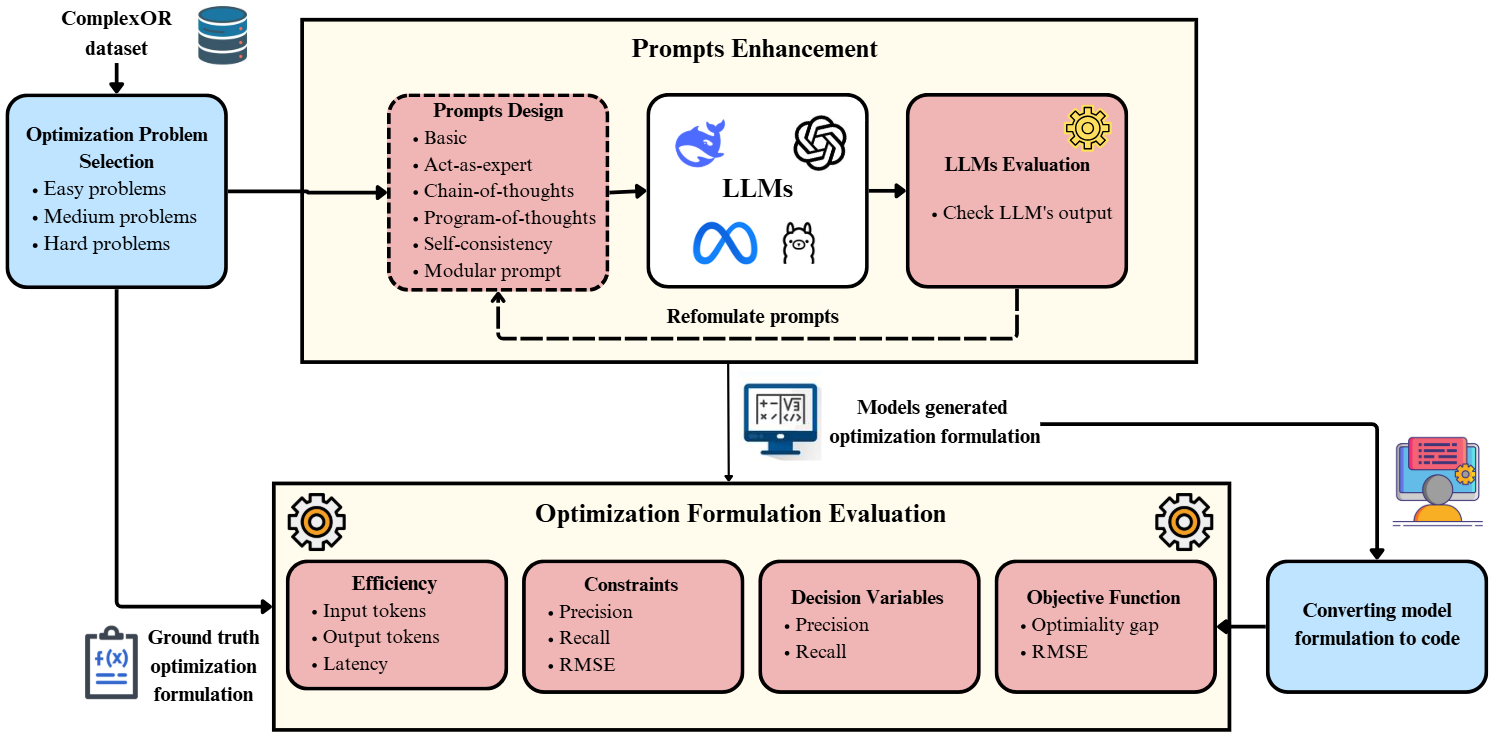}
    \caption{The main phases of the adopted methodology.}
    \label{fig:Methodology}
\end{figure*}

This section presents the experimental setup and methodology used to evaluate the performance of LLMs in optimization modeling. The overall workflow, illustrated in Fig.~\ref{fig:Methodology}, is organized into four main phases.
First, we select a set of benchmark optimization problems with varying levels of complexity to ensure comprehensive coverage across different formulation types and difficulty tiers. Second, we employ multiple prompting strategies to guide the models in generating optimization formulations. The prompt–response loop is iteratively refined until we obtain a final set of prompt templates that yield the most accurate and consistent formulations from the LLMs. Third, we evaluate the generated formulations using the component-level metrics introduced in Section~\ref{sec:evaluationMetrics}, encompassing structural correctness, numerical fidelity, solution quality, and efficiency. Finally, we conduct a comparative analysis across LLMs, prompting strategies, and problem complexities to uncover systematic error patterns and trade-offs in model behavior. The following subsections elaborate on each part of this framework in detail, covering the benchmark problems, prompting strategies, and the evaluated LLMs.


\vspace{-3mm}
\subsection{Optimization Problems}\label{OptimizationProblems}
To evaluate the ability of LLMs to handle optimization tasks of varying difficulty, we selected four representative problems from the ComplexOR dataset~\cite{COMPLEXOR}. These problems differ in structural and computational complexity, enabling us to assess how effectively various prompting techniques guide LLMs in translating natural language descriptions into valid mathematical formulations and solving the corresponding optimization problems. Difficulty levels: Easy, medium, and hard, are assigned based on intrinsic factors such as problem dimensionality, constraint coupling, and variable interactions. An overview of the selected problems is presented in Table~\ref{tab:OptimizationProblems}, while detailed information about problems is provided in~\ref{appendix:problem_details}.

\begin{table}[H]
\centering
\caption{Representative optimization problems categorized by type and complexity.}
\label{tab:OptimizationProblems}
\renewcommand{\arraystretch}{0.9} 
\setlength{\tabcolsep}{4pt}      
\scriptsize                      

\begin{adjustbox}{max width=\linewidth}
\begin{tabular}{|c|l|l|c|p{5.8cm}|}
\hline
\textbf{No.} & \textbf{Problem} & \textbf{Optimization Type} & \textbf{Difficulty} & \textbf{Description} \\ \hline
1 & Knapsack problem~\cite{COMPLEXOR} & Integer linear programming & Easy & Maximize total value of selected items without exceeding the weight limit. \\ \hline
2 & Aircraft assignment problem~\cite{COMPLEXOR} & Mixed-integer linear programming & Medium & Assign aircraft to routes while minimizing total operational costs and meeting demand. \\ \hline
3 & Diet optimization problem~\cite{COMPLEXOR} & Linear programming & Medium & Minimize food costs while satisfying nutritional and dietary requirements. \\ \hline
4 & Aircraft landing problem~\cite{COMPLEXOR} & Mixed-integer linear programming & Hard & Schedule aircraft landings minimizing total delay penalties while respecting separation constraints. \\ \hline
\end{tabular}
\end{adjustbox}
\end{table}


\subsection{Prompting Techniques}\label{PromptingTechniques}

To evaluate how effectively LLMs can translate natural language descriptions of optimization problems into precise mathematical formulations, we employ six prompting strategies that span different levels of guidance and reasoning structure. These strategies were designed to capture a wide spectrum of reasoning depth and structural complexity, ensuring that the evaluation does not depend on a single prompting style. Each is designed to be general-purpose and applicable across diverse optimization problems, regardless of their formulation type or difficulty.

The \textit{basic prompt} provides direct instructions and relies solely on the model’s internal reasoning, while the \textit{act-as-expert prompt} instructs the model to behave like an optimization specialist, systematically identifying objectives and constraints. The \textit{chain-of-thought prompt} encourages explicit step-by-step reasoning to decompose the problem, whereas the \textit{program-of-thought prompt} structures the reasoning in a code-like, algorithmic flow that enhances logical clarity for complex cases. To improve robustness, the \textit{self-consistency prompt} generates multiple reasoning paths and selects the most consistent solution. Finally, the \textit{modular prompting} approach divides the task into sequential stages, with each stage focusing on a specific component, such as decision variables, objectives, or constraints, incrementally building the complete optimization model.  Fig.~\ref{fig:prompting_techniques} presents the six prompting strategies used in this study.


\begin{figure*}[htbp]
\centering

\makebox[\textwidth][c]{%
\resizebox{1.0\linewidth}{!}{%
\begin{tcolorbox}[
  colframe=gray!80!white, colback=white, coltitle=black,
  title=\textbf{Prompting Techniques}, fonttitle=\bfseries,
  boxrule=0.4pt, arc=1.5pt,
  left=2pt, right=2pt, top=2pt, bottom=2pt,
  width=1.45\textwidth, 
  before skip=0mm, after skip=0mm
]

{\small

\tcbset{
  before skip=0mm,
  after skip=0mm,
  top=0pt,
  bottom=0pt,
  left=3pt,
  right=3pt,
  boxrule=0.3pt,
  arc=1pt,
  colframe=purple!60!black,
  colback=purple!7,
  fonttitle=\bfseries,
  coltitle=black
}

\begin{tcolorbox}[colframe=purple!50!black, colback=purple!10, title=\textbf{Prompt 1: Basic Prompt}, boxrule=0.3pt, arc=1pt, top=0pt, bottom=0pt]
Solve the above optimization problem. \vspace{-6mm}
\begin{itemize}
    \item Step 1: Identify the decision variables, objective function, and constraints.\vspace{-6mm}
    \item Step 2: Write the complete mathematical formulation. Do not include any explanation, only output the \vspace{-4mm}
    
    formulation.
\end{itemize}

\end{tcolorbox}

\tcbset{before skip=0mm, after skip=0.8mm}

\begin{tcolorbox}[colframe=purple!50!black, colback=purple!10, title=\textbf{Prompt 2: Act-As-Expert}, boxrule=0.3pt, arc=1pt, top=0pt, bottom=0pt] 
As an expert in optimization, solve the problem by:\vspace{-6mm}
\begin{itemize}
    \item Defining the decision variables.\vspace{-6mm}
    \item Formulating the objective function.\vspace{-6mm}
    \item Listing the constraints.\vspace{-6mm}
\end{itemize}
Then, provide the complete formulation. Do not include any explanation, only output the formulation.
\end{tcolorbox}

\begin{tcolorbox}[colframe=purple!50!black, colback=purple!10, title=\textbf{Prompt 3: Chain-of-Thought}, boxrule=0.3pt, arc=1pt, top=0pt, bottom=0pt]
Solve the problem by thinking step-by-step:\vspace{-6mm}
\begin{itemize}
    \item Identify the objective.\vspace{-6mm}
    \item Define the decision variables.\vspace{-6mm}
    \item Formulate the objective function.\vspace{-6mm}
    \item List the constraints.\vspace{-6mm}
\end{itemize}
Finally, present the full formulation. Do not include any explanation, only output the formulation.
\end{tcolorbox}

\begin{tcolorbox}[colframe=purple!50!black, colback=purple!10, title=\textbf{Prompt 4: Program-of-Thought}, boxrule=0.3pt, arc=1pt, top=0pt, bottom=0pt] 
Use a logic-based reasoning process:\vspace{-6mm}
\begin{itemize}
    \item Define decision variables.\vspace{-6mm}
    \item Specify the objective function.\vspace{-6mm}
    \item Add constraints.\vspace{-6mm}
    \item Identify the model type. Do not include any explanation, only output the formulation.
\end{itemize} 
\end{tcolorbox}

\begin{tcolorbox}[colframe=purple!50!black, colback=purple!10, title=\textbf{Prompt 5: Self-Consistency}, boxrule=0.3pt, arc=1pt, top=0pt, bottom=0pt]  
Generate three formulations independently for the same optimization problem:\vspace{-6mm}
\begin{itemize}
    \item Each should include (decision variables, objective function, and constraints).\vspace{-6mm}
\end{itemize}
Internally compare and select the best one.
Output only the final selected formulation. 
\vspace{-3mm} 

Do not include any explanation or mention of other formulations.
\end{tcolorbox}

\begin{tcolorbox}[colframe=purple!50!black, colback=purple!10, title=\textbf{Prompt 6: Modular Prompting Steps}, boxrule=0.3pt, arc=1pt, top=1pt, bottom=1pt] 
\begin{itemize}
    \item \textbf{Step 1:} Identify and define ONLY the decision variables. Do not solve or explain anything else.
    \vspace{-6mm} 
    \item \textbf{Step 2:} Write ONLY the mathematical expression for the objective function. Do not explain or solve.
    \vspace{-6mm}
    \item \textbf{Step 3:} Write ONLY the mathematical constraints from the problem. Do not solve or interpret.
     \vspace{-6mm}
    \item \textbf{Step 4:} Combine the decision variables, objective function, and constraints into a complete optimization   
    \vspace{-12mm}

    model. Do not explain or solve it.
\end{itemize}
\end{tcolorbox}

} 

\end{tcolorbox}
} 
}
\caption{Prompting techniques for guiding LLMs in optimization formulation.}
\label{fig:prompting_techniques}
\end{figure*}


\subsection{Large Language Models}

In this study, we employed three advanced LLMs for mathematical reasoning and optimization formulation: DeepSeek Math 7B Instruct, fine-tuned for structured mathematical problem-solving; LLaMA 3.1 8B Instruct, Meta’s latest instruction-tuned model with strong reasoning performance; and GPT-5, OpenAI’s newest generation offering state-of-the-art reasoning, efficiency, and robustness for complex formulations. All models were run with a low temperature ($0.1$) to ensure deterministic, consistent outputs. Table~\ref{table:model_details} summarizes their versions, token limits, and accessibility.

\begin{table}[H]
\centering
\caption{LLM models, versions, and configurations used in the experiments.}
\label{table:model_details}
\renewcommand{\arraystretch}{0.9} 
\setlength{\tabcolsep}{4pt}      
\scriptsize                      

\begin{adjustbox}{max width=\linewidth}
\begin{tabular}{c l l c c c}
\toprule
\textbf{No.} & \textbf{Model} & \textbf{Model Version} & \textbf{Temp.} & \textbf{Max Tokens} & \textbf{Source} \\
\midrule
1 & DeepSeek Math & deepseek-math-7b-instruct & 0.1 & 800 & Open \\
2 & LLaMA 3.1 Instruct & meta-llama/Llama-3.1-8B-Instruct & 0.1 & 800 & Open \\
3 & GPT-5 & gpt-5 & 0.1 & 800 & Closed \\
\bottomrule
\end{tabular}
\end{adjustbox}
\end{table}

\vspace{-8mm}
\section{Results and Discussion}\label{sec:discution}
    This section evaluates the performance of LLMs in generating optimization formulations using the proposed evaluation framework. The analysis spans four benchmark problems of varying complexity: Knapsack, aircraft assignment, diet optimization, and aircraft landing, with results summarized in Tables~\ref{table:knapsack_summary}-\ref{table:aircraft_landing_summary}. Each table reports the outcomes of all six prompting strategies across the three selected LLMs, assessed using the complete set of evaluation metrics: Constraint precision and recall, decision variable precision and recall, optimality gap, obj- RMSE, cons-RMSE, and efficiency (latency, input, and output tokens). 

    The discussion is structured in two parts. The first part presents a problem-specific analysis, examining model performance across all metrics within each problem. The second part synthesizes the results across problems, identifying systematic trends, the best-performing models and prompts, and the key relationships between metrics.

\begin{table*}[htbp]
\centering
\caption{Summary of all evaluation metrics for the knapsack problem.}
\label{table:knapsack_summary}
\renewcommand{\arraystretch}{0.9} 
\setlength{\tabcolsep}{2pt}      
\scriptsize                       

\begin{adjustbox}{max width=\textwidth}
\begin{tabular}{c l c c c c c c c c c c}
\toprule
\textbf{Prompt} & \textbf{Model} & \textbf{Cons-P} & \textbf{Cons-R} & \textbf{DV-P} & \textbf{DV-R} & \textbf{Opt. Gap} & \textbf{Obj-RMSE} & \textbf{Cons-RMSE} & \textbf{Latency} & \textbf{Input Tokens} & \textbf{Output Tokens} \\  
\midrule
P1 & DeepSeek & 1.00 & 1.00 & 1.00 & 1.00 & 0.00 & 0.00 & 0.00 & 5.79 & 137 & 205 \\ [-4pt]
   & LLaMA 3.1 & 0.66 & 1.00 & 1.00 & 1.00 & 0.00 & 0.00 & 0.00 & 7.97 & 137 & 251 \\[-4pt]
   & \textbf{GPT-5} & \textbf{1.00} & \textbf{1.00} & \textbf{1.00} & \textbf{1.00} & \textbf{0.00} & \textbf{0.00} & \textbf{0.00} & \textbf{2.00} & \textbf{137} & \textbf{39} \\
\midrule
P2 & DeepSeek & 1.00 & 1.00 & 1.00 & 1.00 & 0.00 & 0.00 & 0.00 & 5.08 & 149 & 186 \\[-4pt]
   & LLaMA 3.1 & 0.66 & 1.00 & 1.00 & 1.00 & 0.00 & 0.00 & 0.00 & 8.12 & 149 & 256 \\[-4pt]
   & \textbf{GPT-5} & \textbf{1.00} & \textbf{1.00} & \textbf{1.00} & \textbf{1.00} & \textbf{0.00} & \textbf{0.00} & \textbf{0.00} & 7.00 & \textbf{149} & \textbf{60} \\
\midrule
P3 & DeepSeek & 1.00 & 1.00 & 1.00 & 1.00 & 0.00 & 0.00 & 0.00 & 1.00 & 158 & 214 \\[-4pt]
   & LLaMA 3.1 & 1.00 & 1.00 & 1.00 & 1.00 & 0.00 & 0.00 & 0.00 & 6.45 & 158 & 204 \\[-4pt]
   & \textbf{GPT-5} & \textbf{1.00} & \textbf{1.00} & \textbf{1.00} & \textbf{1.00} & \textbf{0.00} & \textbf{0.00} & \textbf{0.00} & \textbf{1.00} & \textbf{158} & \textbf{48} \\
\midrule
P4 & DeepSeek & 1.00 & 1.00 & 1.00 & 1.00 & 0.00 & 0.00 & 0.00 & 4.51 & 142 & 164 \\[-4pt]
   & LLaMA 3.1 & 0.66 & 1.00 & 1.00 & 1.00 & 0.00 & 0.00 & 0.00 & 6.41 & 142 & 203 \\[-4pt]
   & \textbf{GPT-5} & \textbf{1.00} & \textbf{1.00} & \textbf{1.00} & \textbf{1.00} & \textbf{0.00} & \textbf{0.00} & \textbf{0.00} & \textbf{1.00} & \textbf{142} & \textbf{50} \\
\midrule
P5 & DeepSeek & 1.00 & 1.00 & 1.00 & 1.00 & 0.00 & 0.00 & 0.00 & 19.66 & 146 & 714 \\[-4pt]
   & LLaMA 3.1 & 1.00 & 1.00 & 1.00 & 1.00 & 0.00 & 0.00 & 0.00 & 6.10 & 146 & 193 \\[-4pt]
   & \textbf{GPT-5} & \textbf{1.00} & \textbf{1.00} & \textbf{1.00} & \textbf{1.00} & \textbf{0.00} & \textbf{0.00} & \textbf{0.00} & 15.00 & \textbf{146} & \textbf{55} \\
\midrule
P6 & DeepSeek & 1.00 & 1.00 & 1.00 & 1.00 & 0.00 & 0.00 & 0.00 & 1.81 & 429 & 60 \\[-4pt]
   & LLaMA 3.1 & 0.66 & 1.00 & 1.00 & 1.00 & 0.00 & 0.00 & 0.00 & 7.32 & 429 & 230 \\[-4pt]
   & \textbf{GPT-5} & \textbf{1.00} & \textbf{1.00} & \textbf{1.00} & \textbf{1.00} & \textbf{0.00} & \textbf{0.00} & \textbf{0.00} & 8.00 & \textbf{429} & \textbf{20} \\
\bottomrule
\end{tabular}
\end{adjustbox}
\end{table*}

\begin{table*}[htbp]
\centering
\caption{Summary of all evaluation metrics for the aircraft assignment problem.}
\label{table:aircraft_assignment_summary}
\renewcommand{\arraystretch}{0.9} 
\setlength{\tabcolsep}{2pt}      
\scriptsize                       

\begin{adjustbox}{max width=\textwidth}
\begin{tabular}{c l c c c c c c c c c c}
\toprule
\textbf{Prompt} & \textbf{Model} & \textbf{Cons-P} & \textbf{Cons-R} & \textbf{DV-P} & \textbf{DV-R} & \textbf{Opt. Gap} & \textbf{Obj-RMSE} & \textbf{Cons-RMSE} & \textbf{Latency} & \textbf{Input Tokens} & \textbf{Output Tokens} \\
\midrule
P1 & DeepSeek & 1.00 & 1.00 & 1.00 & 1.00 & 0.00 & 0.00 & 0.00 & 12.34 & 198 & 439 \\[-4pt]
   & LLaMA 3.1 & 1.00 & 1.00 & 0.50 & 1.00 & 0.04 & 279.00 & 0.00 & 11.32 & 198 & 358 \\[-4pt]
   & \textbf{GPT-5} & \textbf{1.00} & \textbf{1.00} & \textbf{1.00} & \textbf{1.00} & \textbf{0.00} & \textbf{0.00} & \textbf{0.00} & \textbf{11.00} & \textbf{198} & \textbf{159} \\
\midrule
P2 & DeepSeek & 0.66 & 0.66 & 0.50 & 1.00 & 0.00 & 0.00 & 0.07 & 5.91 & 210 & 210 \\[-4pt]
   & LLaMA 3.1 & 1.00 & 1.00 & 0.50 & 1.00 & 0.04 & 263.00 & 0.00 & 11.64 & 210 & 364 \\[-4pt]
   & \textbf{GPT-5} & \textbf{1.00} & \textbf{1.00} & \textbf{1.00} & \textbf{1.00} & \textbf{0.00} & \textbf{0.00} & \textbf{0.00} & 10.00 & 210 & 329 \\
\midrule
P3 & DeepSeek & 0.75 & 1.00 & 0.50 & 1.00 & 0.45 & 525.35 & 0.00 & 9.28 & 219 & 340 \\[-4pt]
   & LLaMA 3.1 & 0.66 & 0.66 & 0.50 & 1.00 & -- & 2031.00 & 0.43 & 10.44 & 219 & 331 \\[-4pt]
   & \textbf{GPT-5} & \textbf{1.00} & \textbf{1.00} & \textbf{1.00} & \textbf{1.00} & \textbf{0.00} & \textbf{0.00} & \textbf{0.00} & 13.00 & \textbf{219} & \textbf{292} \\
\midrule
P4 & DeepSeek & 0.66 & 0.66 & 0.50 & 1.00 & 0.12 & 528.83 & 0.07 & 4.43 & 203 & 163 \\[-4pt]
   & LLaMA 3.1 & 0.66 & 0.66 & 0.33 & 1.00 & -- & 2147.00 & 0.45 & 8.23 & 203 & 262 \\[-4pt]
   & \textbf{GPT-5} & \textbf{1.00} & \textbf{1.00} & \textbf{1.00} & \textbf{1.00} & \textbf{0.00} & \textbf{0.00} & \textbf{0.00} & 5.00 & 203 & 253 \\
\midrule
P5 & \textbf{DeepSeek} & \textbf{1.00} & \textbf{1.00} & \textbf{1.00} & \textbf{1.00} & \textbf{0.00} & \textbf{0.00} & \textbf{0.00} & \textbf{7.73} & \textbf{207} & \textbf{281} \\[-4pt]
   & LLaMA 3.1 & 0.33 & 0.33 & 0.33 & 1.00 & 160.00 & 533.00 & 37.63 & 9.91 & 207 & 312 \\[-4pt]
   & GPT-5 & 1.00 & 1.00 & 1.00 & 1.00 & 0.00 & 0.00 & 0.06 & 21.00 & 207 & 460 \\
\midrule
P6 & DeepSeek & 0.50 & 1.00 & 1.00 & 1.00 & 0.45 & 0.00 & 0.00 & 3.56 & 432 & 123 \\[-4pt]
   & LLaMA 3.1 & 1.00 & 1.00 & 1.00 & 1.00 & 0.00 & 252.00 & 0.00 & 6.80 & 432 & 214 \\[-4pt]
   & \textbf{GPT-5} & \textbf{1.00} & \textbf{1.00} & \textbf{1.00} & \textbf{1.00} & \textbf{0.00} & \textbf{0.00} & \textbf{0.00} & 7.00 & \textbf{432} & \textbf{31} \\
\bottomrule
\end{tabular}
\end{adjustbox}
\end{table*}

\begin{table*}[htbp]
\centering
\caption{Summary of all evaluation metrics for the diet problem.}
\label{table:diet_summary}
\renewcommand{\arraystretch}{0.9} 
\setlength{\tabcolsep}{2pt}      
\scriptsize                       

\begin{adjustbox}{max width=\textwidth}
\begin{tabular}{c l c c c c c c c c c c}
\toprule
\textbf{Prompt} & \textbf{Model} & \textbf{Cons-P} & \textbf{Cons-R} & \textbf{DV-P} & \textbf{DV-R} & \textbf{Opt. Gap} & \textbf{Obj-RMSE} & \textbf{Cons-RMSE} & \textbf{Latency} & \textbf{Input Tokens} & \textbf{Output Tokens} \\
\midrule
P1 & DeepSeek & 1.00 & 1.00 & 1.00 & 1.00 & 0.00 & 0.00 & 0.00 & 8.56 & 210 & 328 \\[-4pt]
   & LLaMA 3.1 & 1.00 & 1.00 & 1.00 & 1.00 & 0.00 & 0.00 & 0.00 & 8.56 & 210 & 271 \\[-4pt]
   & \textbf{GPT-5} & \textbf{1.00} & \textbf{1.00} & \textbf{1.00} & \textbf{1.00} & \textbf{0.00} & \textbf{0.00} & \textbf{0.00} & \textbf{2.00} & \textbf{210} & \textbf{209} \\
\midrule
P2 & DeepSeek & 1.00 & 0.80 & 1.00 & 1.00 & 0.00 & 0.00 & 1.42 & 10.49 & 223 & 271 \\[-4pt]
   & LLaMA 3.1 & 1.00 & 1.00 & 1.00 & 1.00 & 0.00 & 0.00 & 0.00 & 10.49 & 223 & 331 \\[-4pt]
   & \textbf{GPT-5} & \textbf{1.00} & \textbf{1.00} & \textbf{1.00} & \textbf{1.00} & \textbf{0.00} & \textbf{0.00} & \textbf{0.00} & \textbf{3.00} & \textbf{223} & \textbf{214} \\
\midrule
P3 & DeepSeek & 1.00 & 1.00 & 1.00 & 1.00 & 0.00 & 0.00 & 0.00 & 6.37 & 229 & 415 \\[-4pt]
   & LLaMA 3.1 & 1.00 & 1.00 & 1.00 & 1.00 & 0.00 & 0.00 & 0.00 & 6.37 & 229 & 199 \\[-4pt]
   & \textbf{GPT-5} & \textbf{1.00} & \textbf{1.00} & \textbf{1.00} & \textbf{1.00} & \textbf{0.00} & \textbf{0.00} & \textbf{0.00} & \textbf{4.00} & \textbf{229} & \textbf{180} \\
\midrule
P4 & \textbf{DeepSeek} & \textbf{1.00} & \textbf{0.80} & \textbf{1.00} & \textbf{1.00} & \textbf{0.00} & \textbf{0.00} & \textbf{1.42} & 7.58 & 216 & 391 \\[-4pt]
   & LLaMA 3.1 & 0.71 & 1.00 & 0.50 & 1.00 & 0.00 & 0.00 & 0.00 & 7.58 & 216 & 241 \\[-4pt]
   & GPT-5 & 1.00 & 0.40 & 1.00 & 1.00 & 0.00 & 0.00 & 2.90 & 0.60 & 216 & 68 \\
\midrule
P5 & DeepSeek & 1.00 & 0.80 & 1.00 & 1.00 & 0.00 & 0.00 & 1.42 & 19.17 & 219 & 467 \\[-4pt]
   & LLaMA 3.1 & 1.00 & 0.80 & 1.00 & 1.00 & 0.00 & 0.00 & 1.42 & 19.17 & 219 & 610 \\[-4pt]
   & \textbf{GPT-5} & \textbf{1.00} & \textbf{1.00} & \textbf{1.00} & \textbf{1.00} & \textbf{0.00} & \textbf{0.00} & \textbf{0.00} & \textbf{2.00} & \textbf{219} & \textbf{210} \\
\midrule
P6 & DeepSeek & 1.00 & 0.80 & 0.00 & 0.00 & 0.00 & 0.00 & 14.00 & 5.68 & 476 & 297 \\[-4pt]
   & LLaMA 3.1 & 1.00 & 1.00 & 1.00 & 1.00 & 0.00 & 0.00 & 0.00 & 5.68 & 476 & 178 \\[-4pt]
   & \textbf{GPT-5} & \textbf{1.00} & \textbf{1.00} & \textbf{1.00} & \textbf{1.00} & \textbf{0.00} & \textbf{0.00} & \textbf{0.00} & \textbf{0.60} & \textbf{476} & \textbf{34} \\
\bottomrule
\end{tabular}
\end{adjustbox}
\end{table*}

\begin{table*}[htbp]
\centering
\caption{Summary of all evaluation metrics for the aircraft landing problem.}
\label{table:aircraft_landing_summary}
\renewcommand{\arraystretch}{0.9} 
\setlength{\tabcolsep}{2pt}      
\scriptsize                       

\begin{adjustbox}{max width=\textwidth}
\begin{tabular}{c l c c c c c c c c c c}
\toprule
\textbf{Prompt} & \textbf{Model} & \textbf{Cons-P} & \textbf{Cons-R} & \textbf{DV-P} & \textbf{DV-R} & \textbf{Opt. Gap} & \textbf{Obj-RMSE} & \textbf{Cons-RMSE} & \textbf{Latency} & \textbf{Input Tokens} & \textbf{Output Tokens} \\
\midrule
P1 & DeepSeek & 1.00 & 0.66 & 1.00 & 0.75 & 0.00 & 7.78 & 1.65 & 10.75 & 248 & 393 \\[-4pt]
   & LLaMA 3.1 & 1.00 & 0.66 & 0.50 & 0.50 & 0.85 & 86.37 & 1.85 & 13.65 & 248 & 432 \\[-4pt]
   & \textbf{GPT-5} & \textbf{1.00} & \textbf{0.75} & \textbf{1.00} & \textbf{0.75} & \textbf{0.00} & \textbf{0.00} & \textbf{1.57} & 24.00 & 248 & \textbf{248} \\
\midrule
P2 & DeepSeek & 1.00 & 0.75 & 1.00 & 0.25 & 0.24 & 0.00 & 3.22 & 12.16 & 260 & 445 \\[-4pt]
   & LLaMA 3.1 & 0.75 & 0.50 & 0.50 & 0.50 & 0.85 & 86.37 & 2.98 & 16.10 & 260 & 509 \\[-4pt]
   & \textbf{GPT-5} & \textbf{1.00} & \textbf{0.75} & \textbf{1.00} & \textbf{0.75} & \textbf{0.00} & \textbf{0.00} & \textbf{1.57} & 27.00 & 260 & \textbf{247} \\
\midrule
P3 & DeepSeek & 1.00 & 0.50 & 1.00 & 0.25 & 0.89 & 195.57 & 2.91 & 14.26 & 269 & 522 \\[-4pt]
   & LLaMA 3.1 & 0.75 & 0.50 & 0.50 & 0.50 & 0.85 & 86.54 & 2.98 & 15.63 & 269 & 496 \\[-4pt]
   & \textbf{GPT-5} & \textbf{1.00} & \textbf{0.75} & \textbf{1.00} & \textbf{0.75} & \textbf{0.00} & \textbf{0.00} & \textbf{1.57} & 22.00 & 269 & \textbf{161} \\
\midrule
P4 & \textbf{DeepSeek} & \textbf{1.00} & \textbf{1.00} & \textbf{1.00} & 0.25 & \textbf{0.92} & \textbf{0.00} & \textbf{0.00} & 5.85 & 253 & 213 \\[-4pt]
   & LLaMA 3.1 & 1.00 & 1.00 & 0.33 & 0.25 & 0.85 & 102.24 & 0.00 & 10.29 & 253 & 325 \\[-4pt]
   & GPT-5 & 1.00 & 0.75 & 1.00 & 0.75 & 0.00 & 0.00 & 1.57 & 56.00 & 253 & 98 \\
\midrule
P5 & DeepSeek & 1.00 & 0.50 & 1.00 & 0.25 & 1.00 & 258.20 & 1.65 & 20.73 & 257 & 755 \\[-4pt]
   & LLaMA 3.1 & 1.00 & 0.66 & 0.50 & 0.25 & 1.00 & 85.15 & 1.85 & 13.82 & 257 & 435 \\[-4pt]
   & \textbf{GPT-5} & \textbf{1.00} & \textbf{1.00} & \textbf{1.00} & \textbf{0.75} & \textbf{0.00} & \textbf{0.00} & \textbf{0.00} & 15.00 & 257 & \textbf{168} \\
\midrule
P6 & DeepSeek & 1.00 & 1.00 & 1.00 & 0.25 & 0.75 & 0.00 & 0.00 & 5.80 & 1008 & 199 \\[-4pt]
   & LLaMA 3.1 & 1.00 & 1.00 & 0.25 & 0.25 & --- & 99.33 & 1.85 & 9.70 & 1008 & 298 \\[-4pt]
   & \textbf{GPT-5} & \textbf{1.00} & \textbf{1.00} & \textbf{1.00} & \textbf{1.00} & \textbf{0.00} & \textbf{0.00} & \textbf{0.00} & \textbf{5.00} & \textbf{1008} & \textbf{65} \\
\bottomrule
\end{tabular}
\end{adjustbox}
\end{table*}


\subsection{Problem-Specific Performance Analysis}

\subsubsection{Knapsack Problem} 
For knapsack problem, GPT-5 and DeepSeek achieved perfect constraint precision and recall ($1.00$), reproducing the minimal ground-truth constraints exactly. LLaMA~3.1 maintained full recall but often added redundant conditions, such as non-negativity (\(x_i \geq 0\)), reducing its precision to $0.66$. These extra constraints did not affect feasibility but lowered structural fidelity.

All models handled decision variables flawlessly, with precision and recall of $1.00$, indicating no omissions or errors. Solver-level metrics confirmed correctness: the optimality gap and objective RMSE were $0.00$ for every model and prompt, demonstrating exact reproduction of the objective. Constraint behavior, measured by Cons-RMSE, was also perfect ($0.00$) for GPT-5 and DeepSeek. LLaMA~3.1 matched this functional behavior despite reduced precision due to redundant constraints.

Efficiency highlighted differences: GPT-5 produced the most concise outputs, with lower latency and fewer tokens. DeepSeek was moderately efficient, while LLaMA generated longer outputs with higher latency.

\textit{Summary:} All models reliably captured the knapsack problem, achieving perfect recall and zero optimality gap. GPT-5 delivered the exact minimal formulation most efficiently, DeepSeek matched its correctness but was less efficient, and LLaMA preserved feasibility while reducing precision with extra constraints.


\subsubsection{Aircraft Assignment Problem}

The aircraft assignment problem, a medium complexity case that requires binary assignment and resource allocation constraints, revealed a wider variation in model performance than the knapsack problem. GPT-5 consistently achieved perfect constraint precision and recall ($1.00$) across all prompts. DeepSeek performed well but dropped to $0.66$ in Prompts~2 and~4 due to omitted binary restrictions. LLaMA~3.1 was the least stable, ranging from perfect performance to $0.33$--$0.66$ when constraints appeared in auxiliary or non-standard forms.

Decision variable identification mirrored this pattern: GPT-5 maintained perfect precision and recall, while DeepSeek and LLaMA occasionally mis-specified variables, especially under non-standard formulations. On the other hand, solver-level metrics confirmed these trends. GPT-5 achieved an optimality gap of $0.00$ across prompts. DeepSeek showed minor deviations, while LLaMA produced large errors, with objective RMSE exceeding $2000$ in some cases. Cons-RMSE followed a similar pattern: GPT-5 was nearly perfect, DeepSeek deviated slightly in Prompts~2 and~4 (Cons-RMSE $0.07$), and LLaMA showed severe deviations (up to $37.63$) due to non-standard formulations.

Efficiency further distinguished the models. GPT-5 generated concise outputs with low token usage and moderate latency. DeepSeek was less efficient, and LLaMA was the least efficient, producing longer outputs with variable latency.

\textit{Summary:} The aircraft assignment problem exposed larger gaps between models. GPT-5 was the most robust, consistently accurate, and efficient. DeepSeek was generally reliable but occasionally sensitive to prompt design, while LLaMA was unstable and prone to redundant or altered constraints that reduced precision and efficiency.


\subsubsection{Diet Optimization Problem} 

The diet optimization problem, a medium-complexity task with continuous variables and multiple nutrient constraints, showed more variability in constraint precision and recall. GPT-5 consistently achieved perfect scores ($1.00$), reliably representing all food- and nutrient-related constraints. DeepSeek generally performed well but dropped recall to $0.80$ when non-negativity conditions were omitted. LLaMA~3.1 was less consistent: precision fell to $0.71$ in Prompt~4 due to redundant or auxiliary constraints, and recall dropped to $0.80$ in other prompts where constraints were missed.

Decision variable handling was highly accurate across all models. GPT-5 maintained perfect precision and recall, while DeepSeek and LLaMA had minor lapses. Solver-level metrics confirmed robustness, with optimality gap and objective RMSE at $0.00$ for all models.

Constraint behavior, measured by Cons-RMSE, highlighted differences. DeepSeek was perfect in Prompts~1 and~3 (Cons-RMSE $0.00$) but showed moderate deviations ($\sim1.42$) in Prompts~2, 4, and~5 due to omitted non-negativity constraints, and severe errors (Cons-RMSE $14.00$) in Prompt~6 from hallucinated variables. LLaMA aligned perfectly in Prompts~1, 2, 3, 4, and~6 (Cons-RMSE $0.00$) but had $1.42$ in Prompt~5. GPT-5 reproduced all constraints correctly except in Prompt~4 (Cons-RMSE $2.90$), reflecting partial but mostly correct behavior.

Efficiency further distinguished models. GPT-5 produced the most concise outputs with the lowest tokens and latency. DeepSeek was less efficient, particularly in Prompt~5, while LLaMA was the least efficient, consistently generating longer outputs with higher and variable latency.

\textit{Summary:} The diet optimization problem revealed clear model differences. GPT-5 was the most reliable and efficient, with only minor omissions. DeepSeek performed well but occasionally missed non-negativity conditions or introduced hallucinations. LLaMA was the least stable, often producing redundant or structurally inconsistent formulations.


\subsubsection{Aircraft Landing Problem} 

The aircraft landing problem is the most complex benchmark case, combining sequencing decisions, time-window feasibility, and pairwise separation constraints. Constraint precision and recall reflect this challenge. GPT-5 maintained high precision but had lower recall in early prompts, partially enforcing separation rules. DeepSeek had perfect precision, but recall dropped to $0.50$ when separation rules were oversimplified. LLaMA~3.1 was unstable, with recall from $0.50$ to $0.66$ when separation rules were misapplied or reversed. Time-window constraints were generally handled well across models.

Decision variable handling varied. GPT-5 and DeepSeek had perfect precision ($1.00$) but weaker recall, dropping to $0.25$ in some prompts. LLaMA performed worst, with both precision and recall frequently below $0.50$. However, solver-level metrics emphasized these differences. GPT-5 consistently had an optimality gap of $0.00$ and an objective RMSE of $0.00$. DeepSeek and LLaMA showed poor alignment, with gaps up to $1.00$ and RMSE exceeding $285$ and $102.24$, respectively.

Constraint behavior (Cons-RMSE) highlighted model robustness. DeepSeek and LLaMA achieved perfect alignment in Prompts~4 and~6 but lower recall in other prompts. GPT-5 showed partial handling of separations in Prompts~1–4 (Cons-RMSE $1.57$), but Prompts~5 and~6 achieved perfect alignment ($0.00$) using structured separation formulations.

Efficiency further distinguished models. GPT-5 had the lowest latency and token usage, particularly in Prompts~5 and~6. DeepSeek was moderately efficient with variable output lengths, while LLaMA was the least efficient, generating longer outputs with high latency.

\textit{Summary:} The aircraft landing problem exposed major weaknesses in constraint modeling. All models handled time windows well, but pairwise separations were challenging. GPT-5 was the most effective, producing accurate, complete, and efficient formulations. DeepSeek achieved partial success, often relying on simplified rules, while LLaMA was the least consistent, frequently misapplying or oversimplifying separation constraints.

\subsection{General Insights Across Problems}
This section synthesizes insights from the four problems to provide a cross-cutting view of how LLMs perform in generating optimization formulations. We organize the discussion into three parts: (i) Identifying the best-performing model for each problem, (ii) comparing the effectiveness of different prompting strategies, and (iii) analyzing the relationships between evaluation metrics. Together, these perspectives provide a holistic understanding of strengths, weaknesses, and systematic trends across models and prompts.


\subsubsection{Best Model per Problem}
This section presents a cross-problem comparison to evaluate how the three models: GPT-5, DeepSeek, and LLaMA~3.1, perform across different levels of problem complexity. The aim is to capture systematic strengths and weaknesses by jointly considering constraint handling, solver outcomes, and efficiency. 

\begin{figure*}[htbp]
\centering
\begin{tabular}{cc}
\includegraphics[width=0.5\textwidth]{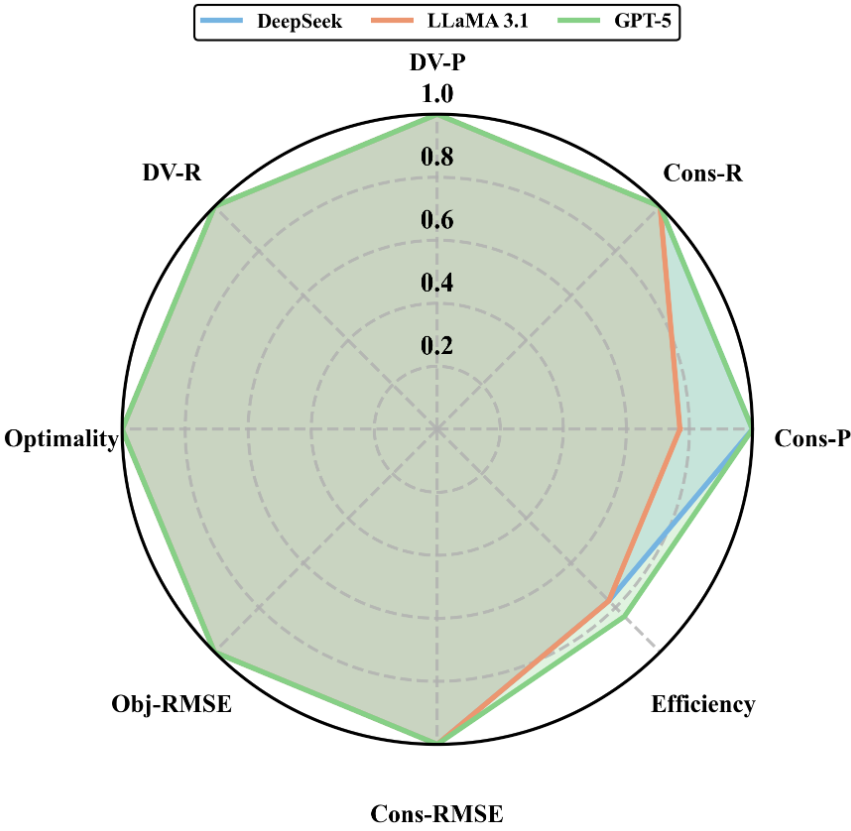} &
\includegraphics[width=0.5\textwidth]{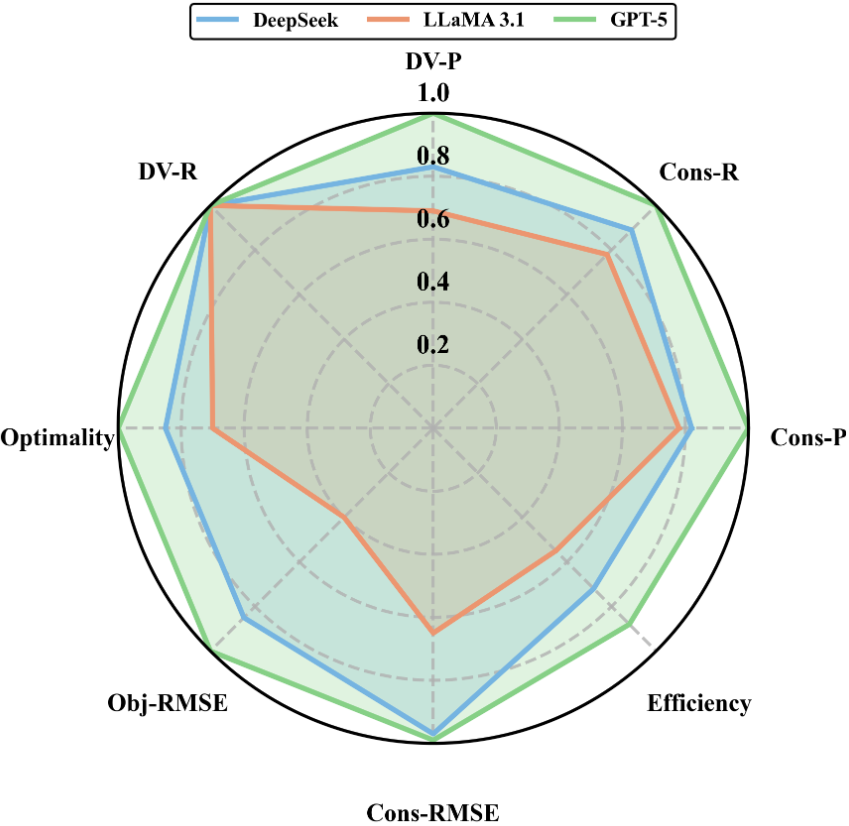} \\
(a) Knapsack problem & (b) Aircraft assignment problem \\ \\
\includegraphics[width=0.5\textwidth]{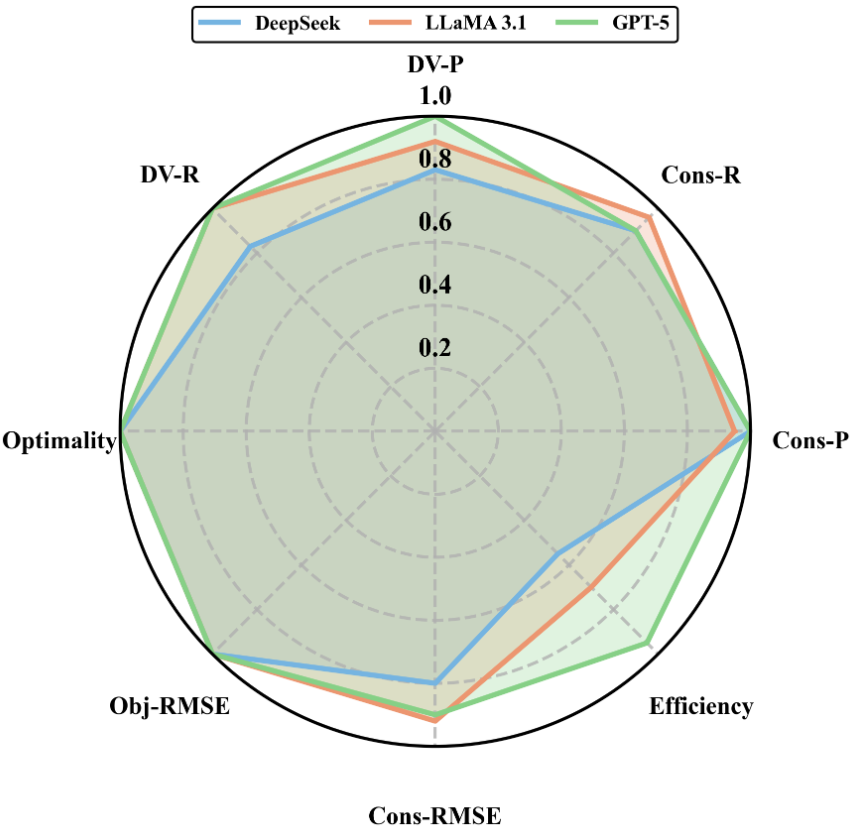} &
\includegraphics[width=0.5\textwidth]{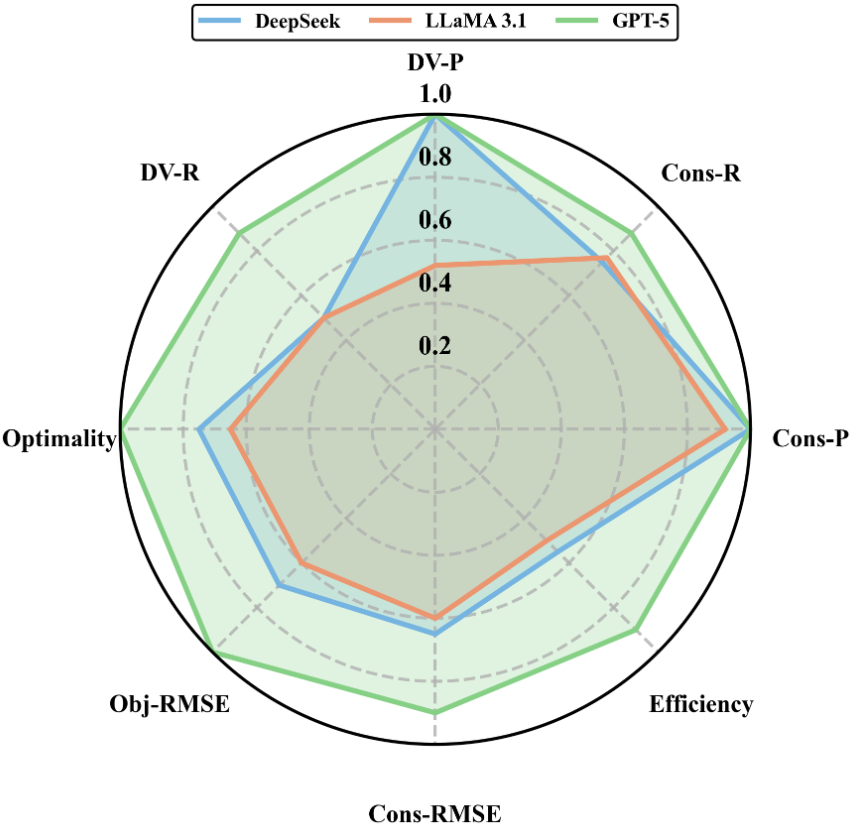} \\
(c) Diet optimization problem & (d) Aircraft landing problem \\
\end{tabular}
\caption{Radar chart comparison of model performance across four benchmark optimization problems. Each axis represents an evaluation metric: Constraint precision (Cons-P), constraint recall (Cons-R), decision variable precision (DV-P), decision variable recall (DV-R), optimality gap (Optimality), objective RMSE (Obj-RMSE), constraint RMSE (Cons-RMSE), and efficiency (Latency, input tokens, output tokens). Higher values indicate better performance.}
\label{fig:radar_all_problems}
\end{figure*}

The radar charts in Fig.~\ref{fig:radar_all_problems} illustrate model performance across the four benchmark problems. In the simple knapsack problem, all models achieved high scores, but GPT-5 stood out with exact structural fidelity and efficient outputs; DeepSeek was reliable but less efficient, while LLaMA~3.1 preserved recall yet lowered precision by adding redundant constraints. In the aircraft assignment problem, differences became clearer: GPT-5 dominated across all metrics, DeepSeek was generally accurate but sometimes omitted binary constraints, and LLaMA lagged with weaker precision, solver alignment, and efficiency. In the diet optimization problem, GPT-5 again delivered near-perfect scores and the most efficient outputs. DeepSeek handled constraints well but often missed non-negativity conditions, and LLaMA was inconsistent, achieving good recall in some prompts but lower precision and solver stability overall. The gap widened in the complex aircraft landing problem, where only GPT-5 consistently produced solver-ready formulations, DeepSeek struggled with recall and solver correctness, and LLaMA performed worst with markedly lower constraint and solver scores. \textit{Overall, GPT-5 was the most robust model across all problems, followed by DeepSeek, while LLaMA showed the least consistency and efficiency.}

To construct the radar charts, performance scores were first averaged across prompts and then normalized for fair comparison. Metrics where higher values represent better performance were scaled to the $[0,1]$ range using min–max normalization, while error-based metrics were inverted before scaling so that higher values consistently indicated better results. Efficiency was computed from latency, input tokens, and output tokens by normalizing, inverting, and averaging them into a single score. Finally, all normalized scores were averaged across prompts to obtain one representative value per model and problem.


\subsubsection{Best Prompting Strategy}

Beyond comparing models, it is equally important to understand how different prompting strategies affect formulation quality. Because each problem type poses unique structural and semantic challenges, the design of prompts plays a crucial role in guiding models to correctly capture objectives, variables, and constraints. This section identifies the most effective prompts for each problem, discusses why they work, and explains how their performance was systematically evaluated.  

\begin{figure}
\centering
\includegraphics[width=0.7\linewidth]{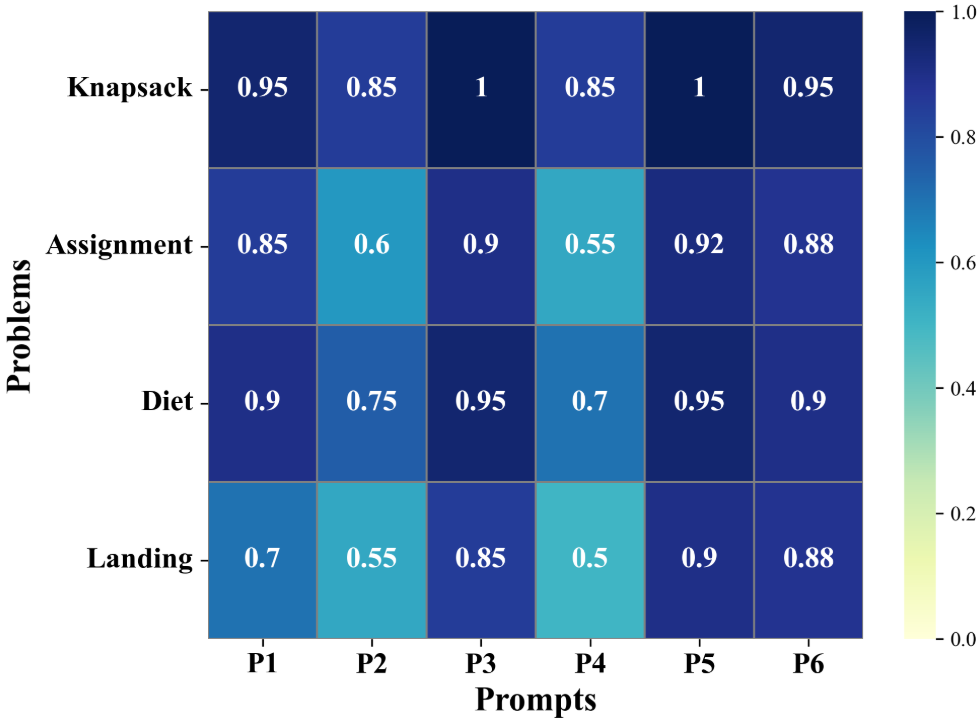}
\caption{Heatmap of prompt effectiveness across problems. Each cell shows the aggregated performance score (0–100\%) for each \emph{problem–prompt} pair. Darker shades indicate better performance, with P3, P5, and P6 showing the highest overall scores.}
\label{fig:BestPrompts}
\end{figure}

As shown in Fig.~\ref{fig:BestPrompts}, prompt effectiveness varied noticeably with problem complexity. In the simple Knapsack problem, Prompts P3 and P5 consistently achieved the highest scores across all models. In the aircraft assignment problem, Prompts P5, P3, P6, and P1 proved most reliable, while P2 and P4 frequently led to missing binary constraints and weaker results. For the diet optimization problem, Prompts P3, P5, P6, and P1 produced the strongest outcomes, whereas P2 and P4 sometimes introduced structural errors. In the most complex aircraft landing problem, Prompts P5 and P6 were essential for achieving full alignment, with P3 also performing strongly, particularly for GPT-5.  

Collectively, \textit{these findings highlight P3 (Chain-of-Thought), P5 (Self-Consistency), and P6 (Modular Prompting Steps) as the most effective strategies overall}. P3 was effective because its structured, step-by-step reasoning reduced logical oversights when identifying objectives, variables, and constraints. P5 improved reliability by generating multiple candidate formulations and internally selecting the best, minimizing omissions and inaccuracies. P6, especially valuable in medium- and high-complexity problems, decomposed the task into sequential stages, first defining variables, then objectives, then constraints, ensuring comprehensive coverage and well-structured outputs. \textit{In particular, P5 and P6 were crucial for complex tasks such as aircraft assignment and aircraft landing, where modular decomposition and redundancy checking were needed to generate solver-ready formulations.}  

To construct the heatmap and evaluate prompt effectiveness across all problems, we computed an aggregate performance score for each \emph{problem–prompt} pair. Metrics where higher values indicate better performance were min–max normalized to $[0,1]$, while error-based and efficiency metrics were inverted before normalization so that higher scores consistently reflected better outcomes. Efficiency (latency, input tokens, output tokens) was averaged into a single score. Scores were then averaged across models and metrics to yield a percentage-like ($0$–$100$\%) measure of overall prompt effectiveness.


\subsubsection{Relationships Between Metrics}

\begin{figure*}[htbp]
\centering
\includegraphics[width=1\linewidth]{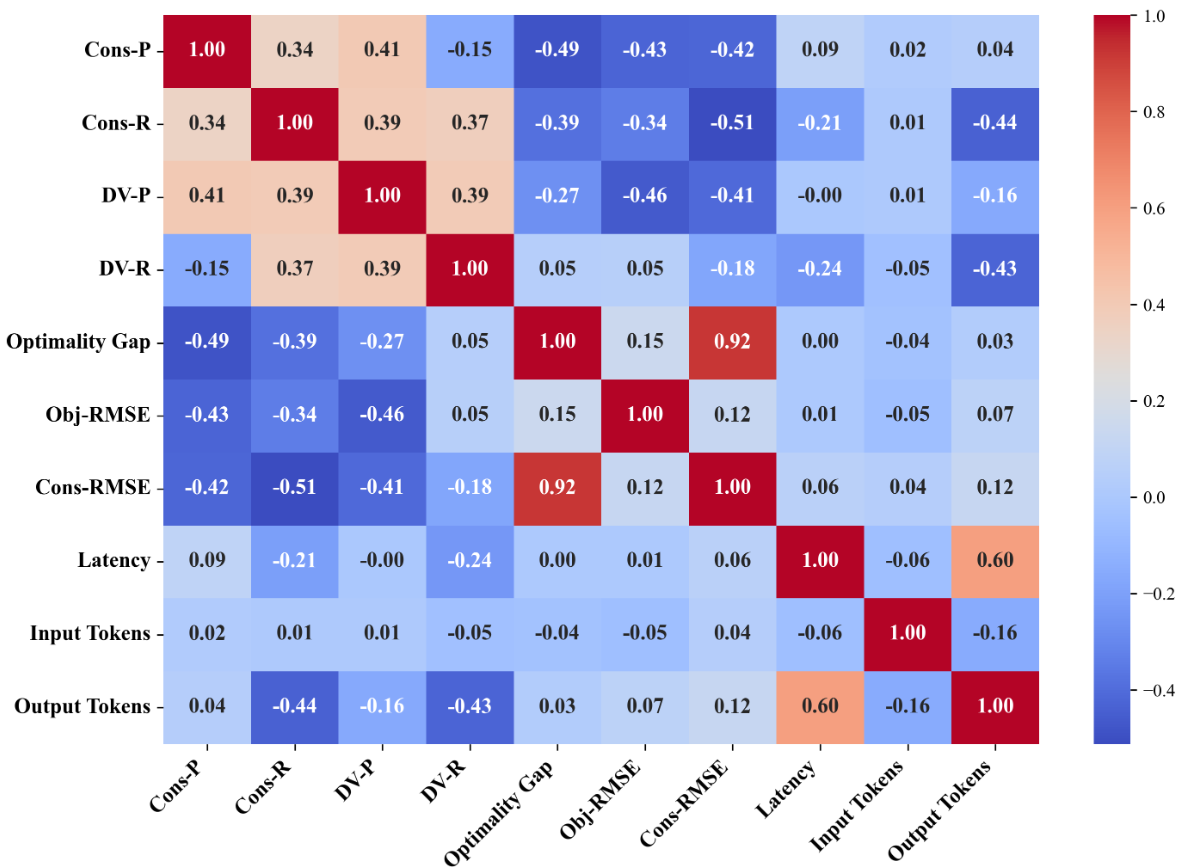}
\caption{Correlation heatmap showing relationships among constraint metrics, solver metrics, and efficiency measures. Red indicates positive and blue negative correlations, with color intensity reflecting strength.}
\label{fig:recall_vs_consrms}
\end{figure*}

To understand how different metrics interact when evaluating generated formulations, we analyze their pairwise correlations. The goal is to identify which metrics most strongly influence solver performance, which ones overlap, and how efficiency relates to correctness. The correlation heatmap in Fig.~\ref{fig:recall_vs_consrms} provides a clear overview of these relationships.

First, \emph{constraint recall (Cons-R)} showed the strongest connection to \emph{constraint RMSE (Cons-RMSE)}. Their correlation was strongly negative ($r=-0.51$), meaning that the more constraints are covered, the fewer violations are observed. When recall = $1$, all constraints are present, and Cons-RMSE is almost always $0$. When recall drops below $1$, at least one constraint is missing: if the missing constraint is critical (e.g., demand satisfaction or aircraft separation), Cons-RMSE rises sharply, but if the missing constraint is redundant (e.g., a non-negativity bound), Cons-RMSE may remain close to $0$. Precision also correlated negatively with Cons-RMSE ($r=-0.42$), but with a weaker effect, confirming that recall is the primary factor for reducing violations. \textit{In short, full recall (Cons-R=$1$) ensures Cons-RMSE is near $0$, while recall below 1 can lead to high Cons-RMSE when critical constraints are missing, though harmless omissions may have little effect.}

Second, \emph{constraint RMSE (Cons-RMSE)} was the most reliable predictor of solver outcomes. Its correlation with the \emph{optimality gap} was extremely strong and positive ($r=0.92$). This means that when Cons-RMSE is close to $0$, indicating that all constraints are satisfied, the optimality gap is also close to $0$, and the solver reaches an optimal solution. By contrast, when Cons-RMSE increases because some constraints are violated, the gap widens sharply, showing that the solution is far from optimal. The correlation with \emph{objective RMSE} was weaker but still positive ($r=0.12$). This shows that constraint violations can sometimes propagate into inconsistencies in the objective values, though this effect is less consistent than for the gap. \textit{In short, when Cons-RMSE is low, the solver produces optimal solutions, but as Cons-RMSE grows, the optimality gap rises and objective values become less reliable.}

Third, the two solver-level metrics behaved differently from each other. The \emph{optimality gap} and \emph{objective RMSE} showed only a weak positive correlation ($r=0.15$), meaning that a large gap does not always imply a large objective error, and vice versa. In practice, the gap is near $0$ when all constraints are satisfied and the solver finds an optimal solution, but it increases when critical constraints are missing. Similarly, Objective RMSE remains close to $0$ when the objective values align with the optimal solution, but rises when constraint violations distort the objective function. Both metrics, however, were moderately negatively correlated with constraint recall (optimality gap: $r=-0.39$, Obj-RMSE: $r=-0.34$), showing that incomplete constraint coverage is the common driver of poor solver performance. \textit{In short, when recall is high, both the gap and objective RMSE stay near $0$; when recall drops, both metrics worsen, even though they are only weakly related to each other.}

Fourth, the constraint metrics were moderately related to decision variable precision but not to decision variable recall. Cons-P and Cons-R were positively correlated with decision variable precision (DV-P: $r=0.41$ with Cons-P, $r=0.39$ with Cons-R), showing that when constraints are represented more precisely and more completely, decision variables also tend to be specified more accurately. By contrast, correlations with decision variable recall (DV-R) were close to $0$, indicating that covering or omitting constraints has little direct influence on whether all decision variables are included. These results suggest that constraint handling mainly improves the correctness of decision variable definitions rather than their completeness. \textit{In short, higher constraint precision and recall both support higher decision variable precision, while decision variable recall remains largely unaffected.}

Finally, efficiency metrics formed a distinct group. \emph{Latency} correlated strongly with \emph{output tokens} ($r=0.60$), showing that more verbose outputs consistently increased inference time. In contrast, Input tokens had near-zero correlations with all other metrics, confirming that efficiency is mainly driven by output verbosity rather than input size. \textit{In practice, when output tokens are large, latency is high, while input length makes little difference.}

\textit{Overall, the heatmap indicates that solver performance depends strongly on two factors: High constraint recall and low cons-RMSE. These metrics directly determine whether the generated formulation satisfies the problem structure and produces correct solutions. Constraint precision and decision variable metrics contribute positively but play a secondary role. Efficiency, in turn, is shaped mainly by output verbosity, with longer outputs increasing latency. In summary, three key principles emerge: Complete constraint coverage prevents violations, minimizing constraint RMSE ensures solver-level correctness, and concise outputs deliver greater efficiency.}


\section{Conclusion and Future Research}\label{sec:Conclusion}

Formulating optimization problems traditionally demands deep domain knowledge, mathematical precision, and iterative collaboration between analysts and engineers, a process that is time-consuming, error-prone, and inaccessible to non-experts. Large language models (LLMs) offer a transformative opportunity to automate this process by translating natural language descriptions into structured mathematical formulations. Yet, despite their advanced reasoning and coding abilities, LLM performance in optimization modeling remains underexplored and lacks systematic, fine-grained evaluation.

We present a comprehensive framework for evaluating LLM-generated optimization formulations at the component level, assessing decision variables, constraints, objective functions, and solution quality. The framework introduces metrics for precision, recall, and root mean squared error (RMSE), complemented by efficiency indicators such as token usage and latency. This multi-dimensional approach provides a diagnostic view of structural correctness, numerical fidelity, and computational efficiency. We validated the framework on four optimization problems of varying complexity using state-of-the-art LLMs (GPT-5, LLaMA, and DeepSeek) and six prompting strategies. Our results identify top-performing models and prompts, revealing trade-offs between structural accuracy, solver quality, and efficiency, and uncover interdependencies among evaluation metrics that shed light on LLM behavior in optimization formulation.

Future work can extend this framework to broader problem sets and additional LLMs, including domain-specialized models, to assess generalizability and architectural impacts on performance. This study lays the foundation for targeted improvements, enabling the development of more reliable and mathematically capable LLMs for optimization problem formulation.


\section*{Acknowledgement}
The authors would like to acknowledge the support received from the Saudi Data and AI Authority (SDAIA) and King Fahd University of Petroleum \& Minerals (KFUPM) under the SDAIA-KFUPM Joint Research Center for Artificial Intelligence Grant JRC-AI-RFP-20. 
\newpage
\appendix

\section{Optimization Problem Details}\label{appendix:problem_details}
This appendix provides detailed descriptions, mathematical formulations, and sample data for the optimization problems used in our experiments. The selected problems include the knapsack problem, aircraft assignment problem, diet problem, and aircraft landing problem, all drawn from the ComplexOR dataset~\cite{COMPLEXOR}.


\begin{figure*}[htbp]
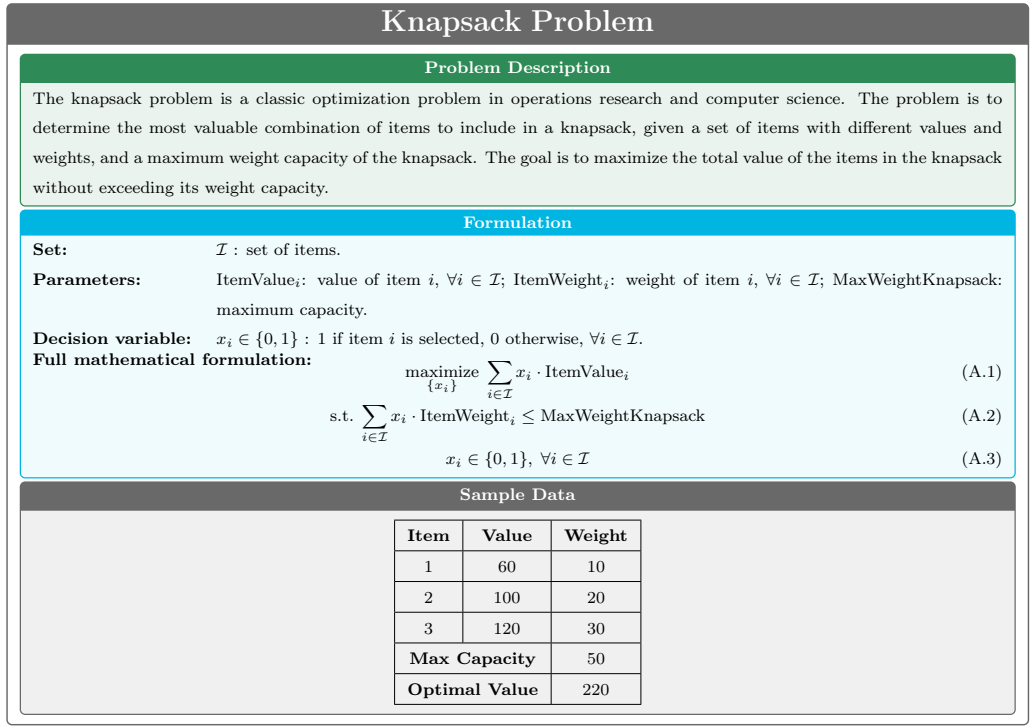

\centering

\definecolor{outer}{HTML}{696969}   
\definecolor{desc}{HTML}{2E8B57}    
\definecolor{form}{HTML}{00B5E2}    
\definecolor{data}{HTML}{696969}    

\makebox[\textwidth][c]{%

\newlength{\probboxw}
\setlength{\probboxw}{1.25\linewidth} 
\resizebox{1.0\linewidth}{!}{%
\begin{tcolorbox}[
    enhanced,
    width=\probboxw,
    colback=white,
    colframe=outer,
    colbacktitle=outer,
    coltitle=white,
    title={\large \textbf{Knapsack Problem}},
    fonttitle=\bfseries,
    halign title=center,
    boxrule=0.7pt,
    arc=2pt,
    top=0.5mm, bottom=0.5mm,
    left=1mm, right=1mm,
    before skip=2mm, after skip=2mm,
    breakable=false
]
\scriptsize

\begin{tcolorbox}[
    width=\linewidth,
    colframe=desc, 
    colback=desc!10, 
    title={\textbf{Problem Description}}, 
    fonttitle=\bfseries,
    halign title=center,
    boxrule=0.6pt,
    arc=2pt,
    left=1mm, right=1mm, top=0.5mm, bottom=0.5mm,
    before skip=0mm, after skip=0.5mm
]
The knapsack problem is a classic optimization problem in operations research and computer science. The problem is to determine the most valuable combination of items to include in a knapsack, given a set of items with different values and weights, and a maximum weight capacity of the knapsack. The goal is to maximize the total value of the items in the knapsack without exceeding its weight capacity.
\end{tcolorbox}

\begin{tcolorbox}[
    width=\linewidth,
    colframe=form, 
    colback=form!6, 
    title={\textbf{Formulation}}, 
    fonttitle=\bfseries,
    halign title=center,
    boxrule=0.6pt,
    arc=2pt,
    left=1mm, right=1mm, top=0.5mm, bottom=0.5mm,
    before skip=0mm, after skip=0.5mm
]
\vspace{-6mm}
\begin{tabularx}{\linewidth}{@{}lX@{}}
\textbf{Set:} & $\mathcal{I}$ : set of items.\\
\textbf{Parameters:} &
$\text{ItemValue}_i$: value of item $i$, $\forall i \in \mathcal{I}$; 
$\text{ItemWeight}_i$: weight of item $i$, $\forall i \in \mathcal{I}$; 
$\text{MaxWeightKnapsack}$: maximum capacity.\\
\textbf{Decision variable:} & 
$x_i \in \{0,1\}$ : 1 if item $i$ is selected, 0 otherwise, $\forall i \in \mathcal{I}$.\\
\end{tabularx}

\noindent\textbf{Full mathematical formulation:}
\vspace{-3mm}
\begin{equation}
\underset{\{x_i\}}{\text{maximize}} \;
\sum_{i \in \mathcal{I}} x_i \cdot \text{ItemValue}_i
\end{equation}
\vspace{-5mm}
\begin{equation}
\text{s.t.} \;
\sum_{i \in \mathcal{I}} x_i \cdot \text{ItemWeight}_i \leq \text{MaxWeightKnapsack}
\end{equation}
\vspace{-5mm}
\begin{equation}
x_i \in \{0, 1\}, \; \forall i \in \mathcal{I}
\end{equation}
\end{tcolorbox}

\begin{tcolorbox}[
    width=\linewidth,
    colframe=data, 
    colback=data!10, 
    title={\textbf{Sample Data}}, 
    fonttitle=\bfseries,
    halign title=center,
    boxrule=0.6pt,
    arc=2pt,
    left=1mm, right=1mm, top=0.5mm, bottom=0.5mm,
    before skip=0mm, after skip=0mm
]
\centering
\scriptsize
\begin{tabular}{|c|c|c|}
\hline
\textbf{Item} & \textbf{Value} & \textbf{Weight} \\ \hline
1 & 60  & 10 \\ \hline
2 & 100 & 20 \\ \hline
3 & 120 & 30 \\ \hline
\multicolumn{2}{|c|}{\textbf{Max Capacity}} & 50 \\ \hline
\multicolumn{2}{|c|}{\textbf{Optimal Value}} & 220 \\ \hline
\end{tabular}
\end{tcolorbox}

\end{tcolorbox}
} 
}
\caption{Illustration of the Knapsack problem, presenting its description, ground-truth mathematical formulation, and sample dataset used in the experimental evaluation.}
\label{fig:Knapsack}
\end{figure*}


\begin{figure*}[htbp]
\centering

\definecolor{outer}{HTML}{696969}   
\definecolor{desc}{HTML}{2E8B57}    
\definecolor{form}{HTML}{00B5E2}    
\definecolor{data}{HTML}{696969}    

\makebox[\textwidth][c]{%

\setlength{\probboxw}{1.26\linewidth} 
\resizebox{1.0\linewidth}{!}{%
\begin{tcolorbox}[
    enhanced,
    width=\probboxw,
    colback=white,
    colframe=outer,
    colbacktitle=outer,
    coltitle=white,
    title={\large \textbf{Aircraft Assignment Problem}},
    fonttitle=\bfseries,
    halign title=center,
    boxrule=0.7pt,
    arc=2pt,
    top=0.5mm, bottom=0.5mm,
    left=1mm, right=1mm,
    before skip=2mm, after skip=2mm,
    breakable=false
]
\scriptsize

\begin{tcolorbox}[
    width=\linewidth,
    colframe=desc, 
    colback=desc!10, 
    title={\textbf{Problem Description}}, 
    fonttitle=\bfseries,
    halign title=center,
    boxrule=0.6pt,
    arc=2pt,
    left=1mm, right=1mm, top=0.5mm, bottom=0.5mm,
    before skip=0mm, after skip=0.5mm
]
The aircraft assignment problem aims to assign aircraft to routes in order to minimize the total cost while satisfying demand constraints with available aircraft. The problem involves a set of aircraft and a set of routes. Given the costs of assigning an aircraft to a route. The objective is to minimize the total cost of the assignment. There are limited available aircraft. It is constrained that the number of each aircraft allocated does not exceed its available number. Given the demand of each route and the capabilities (the largest number of people can be carried) of an aircraft for a route. The demand constraint ensures that the total allocation for each route satisfies the demand. The problem seeks to find the most cost-effective assignment of aircraft to routes.
\end{tcolorbox}

\begin{tcolorbox}[
    width=\linewidth,
    colframe=form, 
    colback=form!6, 
    title={\textbf{Formulation}}, 
    fonttitle=\bfseries,
    halign title=center,
    boxrule=0.6pt,
    arc=2pt,
    left=1mm, right=1mm, top=0.5mm, bottom=0.5mm,
    before skip=0mm, after skip=0.5mm
]
\setlength{\abovedisplayskip}{1pt}
\setlength{\belowdisplayskip}{1pt}
\vspace{-6mm}
\begin{tabularx}{\linewidth}{@{}lX@{}}
\textbf{Sets:} & $\mathcal{A}$ : set of aircraft; $\mathcal{R}$ : set of routes.\\
\textbf{Parameters:} &
$\text{Costs}_{a,r}$: cost of assigning aircraft $a$ to route $r$; 
$\text{Availability}_a$: number of available aircraft of type $a$; 
$\text{Capabilities}_{a,r}$: passenger capacity of aircraft $a$ on route $r$; 
$\text{Demand}_r$: required passengers on route $r$.\\
\textbf{Decision variable:} &
$x_{a,r} \in \{0,1\}$ : 1 if aircraft $a$ is assigned to route $r$, 0 otherwise.\\
\end{tabularx}
\noindent\textbf{Full mathematical formulation:}
\vspace{-2mm}
\begin{equation}
\underset{\{x_{a,r}\}}{\text{minimize}} \;
\sum_{a \in \mathcal{A}} \sum_{r \in \mathcal{R}} \text{Costs}_{a,r} \cdot x_{a,r}
\end{equation}
\begin{equation}
\text{s.t.} \;
\sum_{r \in \mathcal{R}} x_{a,r} \leq \text{Availability}_a, \quad \forall a \in \mathcal{A}
\end{equation}
\begin{equation}
\sum_{a \in \mathcal{A}} x_{a,r} \cdot \text{Capabilities}_{a,r} \geq \text{Demand}_r, \quad \forall r \in \mathcal{R}
\end{equation}
\vspace{-5mm}
\begin{equation}
x_{a,r} \in \{0, 1\}, \quad \forall a \in \mathcal{A}, \; r \in \mathcal{R}
\end{equation}
\end{tcolorbox}

\begin{tcolorbox}[
    width=\linewidth,
    colframe=data, 
    colback=data!10, 
    title={\textbf{Sample Data}}, 
    fonttitle=\bfseries,
    halign title=center,
    boxrule=0.6pt,
    arc=2pt,
    left=1mm, right=1mm, top=0.5mm, bottom=0.5mm,
    before skip=0mm, after skip=0mm
]
\scriptsize
\textbf{Aircraft Availability:} [2, 3, 1], \textbf{Route Demand:} [100, 150]

\textbf{Capabilities Matrix (Passengers per Aircraft per Route):}
\vspace{-3mm}
\begin{center}
\begin{tabular}{|c|c|c|}
\hline
\textbf{Aircraft} & \textbf{Route 1} & \textbf{Route 2} \\ \hline
1 & 50 & 70 \\ \hline
2 & 60 & 80 \\ \hline
3 & 70 & 90 \\ \hline
\end{tabular}
\end{center}
\vspace{-3mm}
\textbf{Cost Matrix (Assignment Costs):}
\vspace{-6mm}
\begin{center}
\begin{tabular}{|c|c|c|}
\hline
\textbf{Aircraft} & \textbf{Route 1} & \textbf{Route 2} \\ \hline
1 & 100 & 200 \\ \hline
2 & 150 & 250 \\ \hline
3 & 200 & 300 \\ \hline
\end{tabular}
\end{center}
\vspace{-3mm}
\textbf{Optimal Solution Value:} 700
\end{tcolorbox}

\end{tcolorbox}
} 
}
\caption{Illustration of the aircraft assignment problem, presenting its description, ground-truth mathematical formulation, and sample dataset used in the experimental evaluation.}
\label{fig:aircraft_assignment_problem}
\end{figure*}


\begin{figure*}[htbp]
\centering

\definecolor{outer}{HTML}{696969}   
\definecolor{desc}{HTML}{2E8B57}    
\definecolor{form}{HTML}{00B5E2}    
\definecolor{data}{HTML}{696969}    

\makebox[\textwidth][c]{%

\setlength{\probboxw}{1.25\linewidth} 
\resizebox{1.0\linewidth}{!}{%
\begin{tcolorbox}[
    enhanced,
    width=\probboxw,
    colback=white,
    colframe=outer,
    colbacktitle=outer,
    coltitle=white,
    title={\large \textbf{Diet Optimization Problem}},
    fonttitle=\bfseries,
    halign title=center,
    boxrule=0.7pt,
    arc=2pt,
    top=0.5mm, bottom=0.5mm,
    left=1mm, right=1mm,
    before skip=2mm, after skip=2mm,
    breakable=false
]
\scriptsize

\begin{tcolorbox}[
    width=\linewidth,
    colframe=desc, 
    colback=desc!10, 
    title={\textbf{Problem Description}}, 
    fonttitle=\bfseries,
    halign title=center,
    boxrule=0.6pt,
    arc=2pt,
    left=1mm, right=1mm, top=0.5mm, bottom=0.5mm,
    before skip=0mm, after skip=0.5mm
]
Consider a diet problem. Given a set of nutrients $\mathcal{N}$ and a set of foods $\mathcal{F}$. 
Each food $j \in \mathcal{F}$ has a cost $Cost_{j}$ and an allowable purchase range $[MinAmount_{j}, \, MaxAmount_{j}]$. 
Each nutrient $i \in \mathcal{N}$ has a required intake range $[MinNutrient_{i}, \, MaxNutrient_{i}]$. 
The amount of nutrient $i$ contained in food $j$ is denoted by $NutrientAmount_{i,j}$. 
The objective is to minimize the total cost of purchased foods, subject to the constraint that the total amount of each nutrient $i$ across all chosen foods lies within its specified range. 
The decision question is: how much of each food $j$ should be purchased?
\end{tcolorbox}

\begin{tcolorbox}[
    width=\linewidth,
    colframe=form, 
    colback=form!6, 
    title={\textbf{Formulation}}, 
    fonttitle=\bfseries,
    halign title=center,
    boxrule=0.6pt,
    arc=2pt,
    left=1mm, right=1mm, top=0.5mm, bottom=0.5mm,
    before skip=0mm, after skip=0.5mm
]
\vspace{-6mm}
\begin{tabularx}{\linewidth}{@{}lX@{}}
\textbf{Sets:} & $\mathcal{F}$ : set of foods; $\mathcal{N}$ : set of nutrients.\\
\textbf{Parameters:} &
$\text{Cost}_j$: cost of food $j$; 
$\text{MinAmount}_j$: minimum amount of food $j$; 
$\text{MaxAmount}_j$: maximum amount of food $j$; 
$\text{NutrientAmount}_{i,j}$: amount of nutrient $i$ in food $j$; 
$\text{MinNutrient}_i$: minimum amount of nutrient $i$ required; 
$\text{MaxNutrient}_i$: maximum amount of nutrient $i$ allowed.\\
\textbf{Decision variable:} &
$x_j$: quantity of food $j$ to include in the diet, $\forall j \in \mathcal{F}$.\\
\end{tabularx}

\noindent\textbf{Full mathematical formulation:}
\vspace{-3mm}
\begin{equation}
\underset{\{x_j\}_{j \in \mathcal{F}}}{\text{minimize}} \;
\sum_{j \in \mathcal{F}} \text{Cost}_j \cdot x_j
\end{equation}
\vspace{-5mm}
\begin{equation}
\text{s.t.} \;
\text{MinAmount}_j \leq x_j \leq \text{MaxAmount}_j, \quad \forall j \in \mathcal{F}
\end{equation}
\vspace{-5mm}
\begin{equation}
\sum_{j \in \mathcal{F}} \text{NutrientAmount}_{i,j} \cdot x_j \geq \text{MinNutrient}_i, \quad \forall i \in \mathcal{N}
\end{equation}
\vspace{-5mm}
\begin{equation}
\sum_{j \in \mathcal{F}} \text{NutrientAmount}_{i,j} \cdot x_j \leq \text{MaxNutrient}_i, \quad \forall i \in \mathcal{N}
\end{equation}
\vspace{-5mm}
\begin{equation}
x_j \geq 0, \quad \forall j \in \mathcal{F}
\end{equation}
\end{tcolorbox}

\begin{tcolorbox}[
    width=\linewidth,
    colframe=data, 
    colback=data!10, 
    title={\textbf{Sample Data}}, 
    fonttitle=\bfseries,
    halign title=center,
    boxrule=0.6pt,
    arc=2pt,
    left=1mm, right=1mm, top=0.5mm, bottom=0.5mm,
    before skip=0mm, after skip=0mm
]
\scriptsize
\textbf{Foods:} Apple, Banana \quad \textbf{Nutrients:} VitaminC, Fiber

\textbf{Food Costs and \\ Bounds:}
\vspace{-11mm}
\begin{center}
\begin{tabular}{|c|c|c|c|}
\hline
\textbf{Food} & \textbf{Cost} & \textbf{Min Amount} & \textbf{Max Amount} \\ \hline
Apple  & 2.0 & 0 & 10 \\ \hline
Banana & 1.5 & 0 & 10 \\ \hline
\end{tabular}
\end{center}
\vspace{-3mm}
\textbf{Nutrient Bounds:}
\vspace{-6mm}
\begin{center}
\begin{tabular}{|c|c|c|}
\hline
\textbf{Nutrient} & \textbf{Min Required} & \textbf{Max Allowed} \\ \hline
VitaminC & 50 & 100 \\ \hline
Fiber    & 30 & 60 \\ \hline
\end{tabular}
\end{center}
\vspace{-2.5mm}
\textbf{Nutrient Amounts \\ (per unit of food):}
\vspace{-12mm}
\begin{center}
\begin{tabular}{|c|c|c|}
\hline
 & \textbf{Apple} & \textbf{Banana} \\ \hline
VitaminC & 10 & 5 \\ \hline
Fiber    & 5 & 10 \\ \hline
\end{tabular}
\end{center}
\vspace{-3mm}
\textbf{Optimal Solution Value:} 10.33
\end{tcolorbox}

\end{tcolorbox}
} 
}
\caption{Illustration of the diet problem, presenting its description, ground-truth mathematical formulation, and sample dataset used in the experimental evaluation.}
\label{fig:diet_problem}
\end{figure*}


\begin{figure*}[htbp]
\centering

\definecolor{outer}{HTML}{696969}   
\definecolor{desc}{HTML}{2E8B57}    
\definecolor{form}{HTML}{00B5E2}    
\definecolor{data}{HTML}{696969}    

\makebox[\textwidth][c]{%
\resizebox{1.0\linewidth}{!}{%
\begin{tcolorbox}[
    enhanced,
    width=1.28\linewidth, 
    colback=white,
    colframe=outer,
    colbacktitle=outer,
    coltitle=white,
    title={\large \textbf{Aircraft Landing Problem (ALP)}},
    fonttitle=\bfseries,
    halign title=center,
    boxrule=0.7pt,
    arc=2pt,
    top=0.5mm, bottom=0.5mm,
    left=1mm, right=1mm,
    before skip=2mm, after skip=2mm,
    breakable=false
]
\scriptsize

\begin{tcolorbox}[
    width=\linewidth,
    colframe=desc, 
    colback=desc!10, 
    title={\textbf{Problem Description}}, 
    fonttitle=\bfseries,
    halign title=center,
    boxrule=0.6pt,
    arc=2pt,
    left=1mm, right=1mm, top=0.5mm, bottom=0.5mm,
    before skip=0mm, after skip=0.5mm
]
The Aircraft Landing Problem (ALP) is the problem of deciding a landing time on an appropriate runway for each aircraft in a given set of aircraft such that each aircraft lands within a predetermined time window; and separation criteria between the landing of an aircraft, and the landing of all successive aircraft, are respected. We are given the earliest landing time, latest landing time, target landing time, and penalties for landing before or after the target landing time for each aircraft. There is also a separation time that represents the minimum time required between the landing of two aircraft. The objective of the problem is to minimize the total penalties of landing before or after the target time for each aircraft. The problem includes several constraints. The order constraint ensures that the aircrafts land in a specific order. The separation constraint ensures that there is enough separation time between the landing of aircraft. The lower and upper time window constraints ensure that each aircraft lands within its respective earliest and latest time windows.
\end{tcolorbox}

\begin{tcolorbox}[
    width=\linewidth,
    colframe=form, 
    colback=form!6, 
    title={\textbf{Formulation}}, 
    fonttitle=\bfseries,
    halign title=center,
    boxrule=0.6pt,
    arc=2pt,
    left=1mm, right=1mm, top=0.5mm, bottom=0.5mm,
    before skip=0mm, after skip=0.5mm
]
\setlength{\abovedisplayskip}{1pt}
\setlength{\belowdisplayskip}{1pt}

\noindent\setlength{\tabcolsep}{0pt}
\setlength{\parindent}{0pt}
\renewcommand{\arraystretch}{0.9}
\vspace{-6mm}
\begin{tabularx}{\linewidth}{@{}lX@{}}
\textbf{Sets and Indices:} & $\mathcal{A}$ : set of aircraft, indexed by $i, j$.\\
\textbf{Parameters:} &
$E_i$: earliest landing time for aircraft $i$; 
$L_i$: latest landing time; 
$T_i$: target landing time; 
$P_i^{\text{early}}$: penalty per unit time if aircraft $i$ lands before $T_i$; 
$P_i^{\text{late}}$: penalty per unit time if aircraft $i$ lands after $T_i$; 
$S_{i,j}$: minimum separation time if $i$ lands before $j$; 
$M$: large constant.\\
\textbf{Decision variables:} &
$x_i$: landing time of aircraft $i$; 
$e_i$: earliness; 
$l_i$: lateness; 
$z_{i,j} \in \{0,1\}$, 1 if $i$ lands before $j$, 0 otherwise.\\
\end{tabularx}
\vspace{-7mm}
\textbf{Full mathematical formulation:}
\vspace{-4mm}
\begin{equation}
\underset{x_i, e_i, l_i, z_{i,j}}{\text{minimize}}
\sum_{i \in \mathcal{A}} P_i^{\text{early}} e_i + P_i^{\text{late}} l_i
\end{equation}
\vspace{-4mm}
\begin{equation}
E_i \leq x_i \leq L_i, \quad \forall i \in \mathcal{A}
\end{equation}
\vspace{-7mm}
\begin{equation}
e_i \geq T_i - x_i, \quad \forall i \in \mathcal{A}
\end{equation}
\vspace{-7mm}
\begin{equation}
l_i \geq x_i - T_i, \quad \forall i \in \mathcal{A}
\end{equation}
\vspace{-7mm}
\begin{equation}
z_{i,j} + z_{j,i} = 1, \quad \forall i \neq j \in \mathcal{A}
\end{equation}
\vspace{-7mm}
\begin{equation}
x_j \geq x_i + S_{i,j} - M(1 - z_{i,j}), \quad \forall i \neq j \in \mathcal{A}
\end{equation}
\end{tcolorbox}

\begin{tcolorbox}[
    width=\linewidth,
    colframe=data, 
    colback=data!10, 
    title={\textbf{Sample Data}}, 
    fonttitle=\bfseries,
    halign title=center,
    boxrule=0.6pt,
    arc=2pt,
    left=1mm, right=1mm, top=0.5mm, bottom=0.5mm,
    before skip=0mm, after skip=0mm
]
\scriptsize
\textbf{Earliest Landing Times:} [1, 3, 5], \textbf{Latest Landing Times:} [10, 12, 15], \textbf{Target Landing Times:} [4, 8, 14]

\textbf{Penalties:}
\vspace{-6mm}
\begin{center}
\begin{tabular}{|c|c|c|}
\hline
\textbf{Aircraft} & \textbf{Early Penalty} & \textbf{Late Penalty} \\ \hline
1 & 5  & 10 \\ \hline
2 & 10 & 20 \\ \hline
3 & 15 & 30 \\ \hline
\end{tabular}
\end{center}
\vspace{-3mm}
\textbf{Separation Times:}
\vspace{-6mm}
\begin{center}
\begin{tabular}{|c|c|c|c|}
\hline
 & \textbf{A1} & \textbf{A2} & \textbf{A3} \\ \hline
A1 & 0 & 2 & 3 \\ \hline
A2 & 2 & 0 & 4 \\ \hline
A3 & 3 & 4 & 0 \\ \hline
\end{tabular}
\end{center}
\vspace{-3mm}
\textbf{Optimal Solution Value:} 0
\end{tcolorbox}

\end{tcolorbox}
} 
}
\caption{Illustration of the aircraft landing problem, presenting its description, ground-truth mathematical formulation, and sample dataset used in the experimental evaluation.}
\label{fig:Airlanding_problem}
\end{figure*}

\newpage
\bibliographystyle{elsarticle-num}
\bibliography{references}

\end{document}